**Can large language models interpret unstructured chat data on dynamic group decision-making processes? Evidence on joint destination choice**

**Sung-Yoo Lim**
Department of Data/Network/AI Convergence
Ajou University, Suwon, Republic of Korea, 16499
Email: syl9205@ajou.ac.kr

**Koki Sato, Co-first Author**
Department of Urban Engineering
The University of Tokyo, Bunkyo-ku, Japan, 113-8656
Email: kk-sato@ut.t.u-tokyo.ac.jp

**Kiyoshi Takami, Ph.D.**
Department of Urban Engineering
The University of Tokyo, Bunkyo-ku, Japan, 113-8656
Email: takami@ut.t.u-tokyo.ac.jp

**Giancarlos Parady, Ph.D.**
Department of Urban Engineering
The University of Tokyo, Bunkyo-ku, Japan, 113-8656
Email: gtroncoso@ut.t.u-tokyo.ac.jp

**Eui-Jin Kim, Ph.D., Corresponding Author**
Department of Transportation Systems Engineering
Ajou University, Suwon, Republic of Korea, 16499
Email: euijin@ajou.ac.kr

**Abstract**
Social activities result from complex joint activity-travel decisions between group members. While observing the decision-making process of these activities is difficult via traditional travel surveys, the advent of new types of data, such as unstructured chat data, can help shed some light on these complex processes. However, interpreting these decision-making processes requires inferring both explicit and implicit factors. This typically involves the labor-intensive task of manually reading and annotating dialogues to capture context-dependent meanings shaped by the social and cultural norms. Against this background, this study evaluates the potential of Large Language Models (LLMs) to automate and complement human annotation in interpreting decision-making processes from group chats, using data on joint eating-out activities in Japan as a case study. We designed a prompting framework inspired by the knowledge acquisition process in Knowledge Graph (KG) construction. This framework sequentially extracts key decision-making factors, including the group-level restaurant choice set and outcome, individual preferences of each alternative, and the specific attributes driving those preferences. This structured process guides the LLM to interpret group chat data step-by-step, converting unstructured dialogues into structured tabular data describing decision-making factors. To evaluate LLM-driven outputs, we conduct a quantitative analysis using a human-annotated ground truth dataset and a qualitative error analysis to examine model limitations. Results show that while the LLM reliably captures explicit decision-making factors, it struggles to identify nuanced implicit factors that human annotators readily identified. We further pinpoint specific contexts where the LLM performs poorly, establishing a boundary for when LLM-based extraction can be trusted versus when human oversight remains essential. These findings highlight both the potential and limitations of LLM-based analysis for incorporating non-traditional data sources on social activities into activity-based models.

**Keywords:** Large language models, Prompt Engineering, Knowledge acquisition, Social activities, Joint decision-making, Leisure travel behavior, Activity-based models, Group chat data.

# 1. Introduction

Human travel choices are strongly influenced by social networks (Axhausen, 2005). In particular, joint travel decisions related to social activities are made in coordination with members of a group, which involves a dynamic social negotiation process (Arentze, 2015a) considering the preferences of each member. Recent studies on joint travel decisions have incorporated heterogeneity in preferences and travel time into individual utility-based models, accounting for the influence of others on individual decisions (Arentze, 2015b; Gramsch-Calvo and Axhausen, 2024a; Han et al., 2023). However, to interpret complex group decision-making processes in detail, it is essential to observe how negotiations actually unfold, that is, how group members negotiate and reach consensus. This requires richer datasets than what is available via traditional travel behavior surveys.

The increasing reliance on ICT has made new data forms available. For example, group chats on messaging platforms such as WhatsApp provide empirical data capturing these complex negotiation processes. This study utilizes group chat data collected via LINE, Japan's most widely used messaging platform, via x-GDP (Text-aided Group Decision-making Process Observation Method), a novel data collection framework (Parady et al., 2025b), to analyze decision-making processes of joint eating-out activities. Such data provides a novel opportunity to incorporate social behavior into activity-based models, by capturing the negotiation and consensus-building processes, which are often overlooked in traditional datasets.

To use group chats for systematic analysis, however, the information in dialogues must be extracted and recorded in a structured format with consistent definitions. As this extraction must reflect domain-specific theories of group decision-making, it requires a methodological framework that specifies what to identify and how to represent it consistently. Traditionally, this process has been performed manually, which is time-consuming and can lead to variability in coding decisions across coders (O'Connor and Joffe, 2020). Knowledge graphs (KGs) provide a robust way to these challenges by representing structured information as entities and relations under a shared schema (Hogan et al., 2022). While KGs provide a powerful representation, the practical challenge lies in automating their construction from unstructured text to ensure consistency and scalability (Ling et al., 2026). Consequently, in this study, we focus on knowledge acquisition, the step immediately preceding KG construction, because analyzing complex joint decision-making processes first requires extracting the essential decision elements and the relations emerging from multi-party conversational interactions.

This knowledge acquisition is challenging as it requires inferring nuanced, implicit information shaped by social and cultural contexts. It involves identifying the choice outcome and the decision-making factors, such as the alternatives within a choice set, the individual characteristics, and the underlying rationale including each individual's preferences for alternative-specific attributes (e.g., restaurant service quality, accessibility, travel times). Conventionally, this demands that researchers manually read and annotate all dialogues to capture implicit, context-dependent information (Goyanes et al., 2025; Lee et al., 2024), a task that is both time- and cost-intensive.

Recent state-of-the-art large language models (LLMs) offer the potential to automate this knowledge acquisition task by extracting explicit and implicit signals from dialogues; however, evidence supporting this potential remains limited. While benchmark studies indicate LLMs equal or surpass human annotators in relatively simple tasks like sentiment labeling, their reliability declines when identifying nuanced implicit cues such as sarcasm (Bojić et al., 2025) or when inferring others' actual beliefs, intentions, or unspoken mental states (Strachan et al., 2024). Therefore, it remains unclear whether LLMs can accurately interpret subtle social dynamics and contextual nuances inherent in complex situations such as joint travel decision-making.

This study aims to evaluate whether an LLM can effectively interpret group chat data, specifically, by converting unstructured dialogues into structured tabular data describing decision-making factors, for joint restaurant choices. In particular, we evaluate whether an LLM can support this knowledge acquisition step by extracting domain-specific entities and relations from group chats through a transparent, stepwise approach. Furthermore, as LLM performance is highly sensitive to prompts, the process of designing and optimizing these inputs (i.e., prompt engineering) must be well-structured to obtain reliable outputs. This study introduces a theory-driven analysis framework for group decision-making, incorporating prompt engineering strategies designed to guide LLMs toward context-appropriate reasoning processes. We conduct (i) a quantitative analysis using an extensive human-annotated ground truth dataset and (ii) a qualitative error analysis to identify specific situations where LLM underperforms.

From unstructured group chat data, our analysis framework extracts both explicit and implicit group decision-making factors, including decision makers, alternatives, and alternative-specific attributes (Bellini-Leite, 2024). These factors are essential for utility-based models, a cornerstone of activity-based modeling with social networks (Gramsch-Calvo et al., 2025; Han et al., 2023; Jin et al., 2025). Understanding joint decision-making contexts requires recognizing two key attributes that distinguish joint choice from individual choice. First, individual egocentrism, where individuals anchor their evaluations on their own preferences while simultaneously adjusting for other group members, plays a crucial role (Arentze, 2015b). Second, final group decisions reflect the collective structure of individual preferences (Chiclana et al., 2007). Therefore, when individuals consistently express the same preference, demonstrating strong egocentrism, they are more likely to influence the final decision, as their clear opinion can shape the group's consensus (Chiclana et al., 2008). These theoretical insights inform our framework design, guiding it to capture both individual preference formation and group preference aggregation.

To operationalize this guidance, we adopt established information extraction concepts from natural language processing (NLP) (Zhong et al., 2024). Specifically, our framework decomposes knowledge acquisition into three tasks: *Entity Discovery* for identifying decision-makers and alternatives, *Coreference Resolution* for tracking repeated entities throughout the conversation, and *Relation Extraction* for capturing alternative-specific attributes discussed during group decision-making. This task decomposition guides the LLM on what to extract at each step, enabling a more systematic and transparent extraction process (Ling et al., 2026).

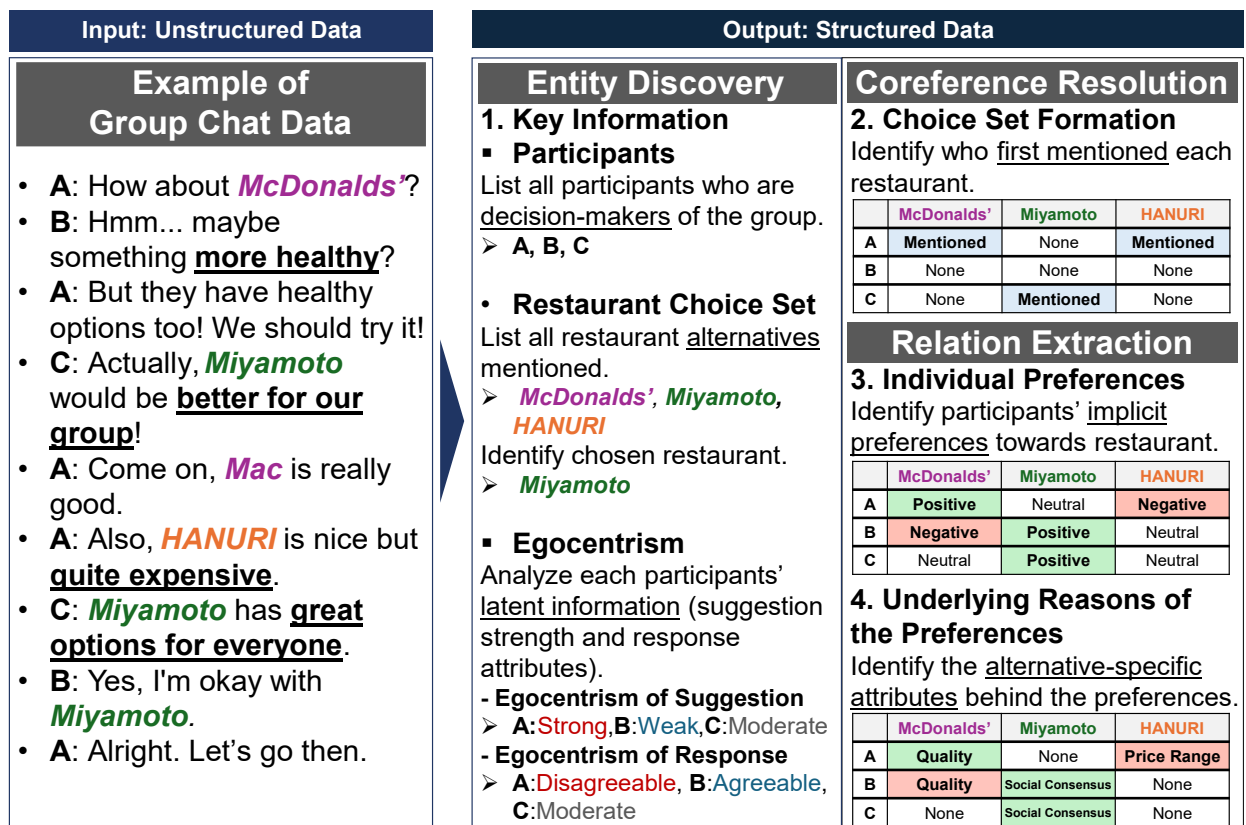

Figure 1 Knowledge acquisition framework: from unstructured group chat to structured entities and relations describing decision-making factors

Building on this methodological foundation, our framework operates through four sequential steps, as shown in **Figure 1**. In the first step, *Entity Discovery* identifies key dialogue information, including the list of participants (decision-makers) and their degree of egocentrism, represented by how participants suggest and respond to one another, as well as the group's restaurant choice set and the final group decision. In the second step, *Coreference Resolution* traces the formation of the choice set (restaurant alternatives) by identifying which participant initially proposed each restaurant amid repeated references throughout the conversation. In the final two steps, *Relation Extraction* captures the relationships between participants and alternatives. Specifically, the third step extracts individual preferences toward each restaurant alternative, while the fourth step identifies the underlying alternative-specific attributes driving these preferences. This sequential structure systematically extracts decision-making factors, revealing how individual preferences are shaped and aggregated into group decisions through social interactions.

While this framework defines the specific knowledge to be extracted, prompt engineering grounded in Dual Process Theory (DPT) (Kahneman, 2011) guides the LLM to utilize appropriate reasoning modes. *Fast thinking* applies to initial information extraction (Step 1) to establish the basic units for subsequent analysis. *Slow thinking*, which supports more deliberative reasoning, applies to subsequent multi-dimensional decision analysis that integrates evidence across participants, alternatives, and their relationships (Steps 2-4). We compare several state-of-the-art prompting strategies to assess how different approaches affect the framework's performance in capturing group decision-making factors.

In summary, this paper explores the potential of LLM-based analysis to understand joint activity-travel decisions using unstructured group chat data, aiming to enhance the flexibility of activity-based models in representing complex social behaviors. Specifically, this study (a) develops a structured framework to interpret unstructured group chat data using LLMs, (b) evaluates its performance in extracting decision-making factors, (c) compares different prompting methods to identify effective designs, and (d) analyzes situations where the LLM struggles to interpret nuanced contextual information.

## 2. Literature Review

This section provides the theoretical and methodological background for our study on group decision-making using LLMs and group chat data. It is organized into four subsections: (a) theoretical background on how individual preferences form joint choices and the role of social interactions, (b) empirical studies comparing LLMs and human experts in interpreting implicit social contexts, (c) knowledge acquisition from unstructured text for building consistent structured representations, and (d) prompt engineering techniques grounded in psychological theory to distinguish between *fast thinking* and *slow thinking* in analyzing group decisions.

### 2.1. Joint Decision-Making in Travel Behavior

Traditional travel behavior models focus on the individual as the primary decision maker, despite widespread evidence that many travel-related decisions are made in coordination with others. Earlier studies incorporating social interactions focused on household-level interactions. Molin et al. (2002) and Zhang and Fujiwara (2009) estimated models considering group-level residential preferences rather than ego-level and showed that incorporating group-level preferences into the model allows for a more accurate representation of household residential choice. Srinivasan and Bhat (2005) focused on maintenance activity participation choices and offered several insights, such as the impact of resource constraints on joint activity participation and the gender-based imbalance in household maintenance burdens.

Focusing on extra-household social interactions, Arentze and Timmermans (2008) proposed a microsimulation of activity patterns that dynamically considered the formation and decline of social networks. Using concepts of the activity based modeling and social networks, Ronald et al. (2012) extended this work by proposing an experimental model that incorporates interpersonal negotiations regarding the type, purpose, location, and timing of activities. However, while these studies provide an agent-based modeling framework to account for social interactions, the required model parameters have yet to be empirically estimated at scale.

From a discrete choice modeling perspective, specifically related to destination choice, Arentze (2015b) developed a stated preference (SP) survey to account for the impact of social network members. They introduced a social utility function that accounts not only for individual preferences but also for those of other group members, explicitly considering equity. The results indicate that individuals' attitudes vary across the decision context. Jin et al. (2025) proposed a novel SP experiment revealing the influence of recognizing friend's preference to individual choice. Their results indicate that group members' preferences and convenience significantly affect individuals' utility, with the magnitude varying by social network attributes and sociodemographics.

From a revealed preference (RP) perspective, Han et al. (2023) estimated discrete choice models incorporating group-level utility for the first time, revealing that the models incorporating group-level impedance outperform ego-only models by up to 32.2%. More recent work by Okamura et al. (2024), Gramsch-Calvo and Axhausen (2024b), and Parady et al. (2025a) also estimated group discrete choice models using RP data, further validating the superior performance of models that incorporate group-level information. Gramsch-Calvo et al. (2025) proposed the concept of "willingness to travel" to capture the tradeoffs inherent in joint decision-making, defined as the marginal rate of substitution between an individual's travel time and the median travel time of the group. Expanding on Neutens et al. (2008), Parady et al. (2025a) empirically estimated models of joint accessibility, explicitly capturing the impact of group travel times and costs, and the available time window while accounting for all members' spatiotemporal constraints.

These studies demonstrate the evolution of research in capturing individual preferences, constraints, and their aggregation into group decisions. Building on this, our study introduces a novel approach that leverages group chat data to examine how individual preferences emerge and converge into group choices through natural dialogue data.

## 2.2 Natural Language Interpretation with LLMs

The analysis of social interactions has traditionally relied on manual human annotation. For decades, researchers have hand-coded natural language data to understand social behavior. In communication studies, emotions and topics in interview transcripts have been manually categorized to identify patterns in words, themes, and concepts (Bengtsson, 2016). In psycholinguistics, researchers have developed word-emotion category mappings to investigate the mental processes underlying language use. In conversation analysis, the precise timing of interaction patterns such as turn-taking and the shifting of speaking roles has been manually annotated from recorded dialogues (Sacks et al., 1974).

Recent advances in LLMs have raised questions about their potential to automate or supplement annotation processes that have traditionally relied on human effort. However, systematic quantitative evaluations show mixed results: while LLMs match or outperform humans on some tasks, they still face challenges with others. Bojić et al. (2025) conducted a comprehensive comparison of human annotators and eight LLMs across four dimensions of content analysis: sentiment, political leaning, emotional intensity, and sarcasm detection. Their findings show that LLMs perform on par with humans in sentiment analysis and surpass human consistency in detecting political leaning, identifying whether a given text expresses liberal, conservative, or other ideological positions. However, LLMs struggle with sarcasm detection, demonstrating low performance and highlighting the inherent difficulty of interpreting implicit and context-dependent meanings.

Beyond quantitative evaluations, studies examining how and why LLMs fail have revealed systematic limitations in their ability to interpret social meaning, particularly in relation to Theory of Mind (ToM), that is, the capacity to understand that others have thoughts, beliefs, and emotions different from one's own (Strachan et al., 2024). Strachan et al. (2024) evaluated GPT-4, GPT-3.5, and LLaMA2 against human participants using five different ToM tests and found that, although GPT-4 performed comparably to humans on most tasks, all models struggled to interpret more complex social situations.

These limitations are of particular importance for applications involving the interpretation of joint decision-making processes, where multiple participants negotiate and reach consensus. Prior research highlights the continued need for human expertise in analyzing group social interactions and underscores the importance of comprehensive evaluation approaches. Such approaches must not only compare quantitative performance between LLMs and human experts but also qualitatively analyze errors to identify where and why LLMs fall short in interpreting social meaning.

## 2.3. LLM-based Knowledge Acquisition from Unstructured Data

Knowledge graphs (KGs) are structured representations of facts, consisting of entities, relationships, and semantic descriptions (Zhong et al., 2024). Hogan et al. (2022) emphasized that KGs are essential for structuring and connecting knowledge to support complex reasoning and decision-making. For a systematic analysis of group chats, decision-making factors must also be represented in a consistent structure to allow for cross-conversation comparison; KGs provide a natural template for this purpose. This capability is particularly relevant for analyzing group decision-making, where extracting who proposed which alternative and how each participant expressed their preferences requires linking information dispersed across a dialogue.

Zhong et al. (2024) identified three stages in KG construction from unstructured data: knowledge acquisition, knowledge refinement, and knowledge evolution. Knowledge acquisition extracts initial information from raw text, while knowledge refinement revises the initial KG to improve its completeness. Knowledge evolution subsequently updates the KG to integrate new facts and evolving patterns over time.

Among these stages, knowledge acquisition serves as the foundation. Ling et al. (2026) noted that errors at this initial stage propagate downstream, compromising the quality of all subsequent refinement and evolution processes. This sequential dependency highlights the critical importance of accurate initial extraction. Accordingly, our study focuses exclusively on knowledge acquisition, as our goal is to extract domain-relevant entities and relations from group chats to serve as consistent structured inputs for analysis.

Knowledge acquisition is commonly operationalized through a set of information extraction subtasks. *Entity discovery* identifies key objects such as people, locations, and concepts, representing the first step in structuring unstructured data (Li et al., 2022). *Coreference resolution* links different expressions that refer to the same entity throughout a text, ensuring consistency and avoiding redundancy (Ling et al., 2026). Finally, *Relation extraction* identifies relationships between entities, including implicit relationships that are not explicitly stated (Nasar et al., 2022).

Ling et al. (2026) demonstrated that LLMs can address challenges in KG construction, such as data scarcity and semantic ambiguity, through their contextual understanding. They categorized LLM-based methods into four approaches: prompt-based, fine-tuning-based, single LLM agent, and multiple LLM agent methods. Among these, prompt-based methods use LLMs as end-to-end models governed by carefully designed prompt templates, guiding the model through step-by-step reasoning.

Building on this framework, we adopt a prompt-based approach to extract decision-making factors from group chat data. By decomposing the task into sequential subtasks (*Entity discovery*, *Coreference resolution*, and *Relation extraction*) with domain-specific prompt instructions, our framework provides structured guidance for capturing the implicit information embedded in group conversations.

### 2.4. Prompt Engineering Design

Prompt engineering, the design of input prompts, has emerged the primary method for optimizing LLM performance without modifying the underlying model parameters. The effect of such prompt design can be understood through Kahneman's Dual Process Theory (DPT) (Bellini-Leite, 2024; He et al., 2024; Zhang et al., 2025), which classifies human thinking into two modes: fast, intuitive cognition (i.e., *fast thinking*) and slow, deliberative reasoning (i.e., *slow thinking*). Just as humans dynamically switch the way of thinking between *fast thinking* and *slow thinking* based on task complexity, prompt design can steer an LLM toward either intuitive pattern recognition or reflective, multi-step reasoning.

*2.4.1. Simple Prompts: Eliciting Fast Thinking*
For tasks amenable to *fast thinking*, where intuitive understanding is sufficient, simple prompts are foten more appropriate than those requiring complex reasoning chains. We evaluate two prompting techniques:
- No-Delimiters (ND): Prompts composed of plain sentences without any structural formatting.
- Zero-Shot (ZS): Prompts that include structural delimiters to define the task but provide no examples or explicit reasoning steps (Kojima et al., 2022).

*2.4.2. Reasoning Prompts: Structuring for Slow Thinking*
Complex tasks that necessitate step-by-step logic or insight into contextual meanings require *slow thinking*. In such cases, simple instruction-based prompts are often insufficient. Prior research has shown that prompts incorporating logical structure and encouraging intermediate reasoning steps significantly improve model accuracy, especially for high-order cognitive tasks. We employ four reasoning-based prompting techniques:

- Chain-of-Thought (CoT): Prompts that guide the model through intermediate inference steps (Wei et al., 2022).
- Self-Refinement (SR): Prompts that instruct the model to critique and revise its initial answer to improve logical consistency (Madaan et al., 2023).
- Prompt Decomposition (PD): Prompts that break down a complex task into simpler subtasks to reduce cognitive load and improve performance (Khot et al., 2022).
- Mixture of Reasoning Experts (MoRE): Prompts that are designed to integrate perspectives from domain-specific experts (Si et al., 2023). We adapt this approach by treating each group participant as an individual "expert," guiding the model to simulate multiple perspectives before aggregating a final conclusion.

## 3. Method
### 3.1. Data

The x-GDP dataset used in this study comprises 47 LINE messenger group chats documenting real-time restaurant choice negotiations (Parady et al., 2025b). This dataset was collected under IRB-approved protocols, featuring real groups of three to five friends coordinating an eating-out activity and choosing a venue, under naturalistic experiment conditions. Each conversation captures the complete decision-making sequence, from initial proposals to preference negotiations and final restaurant selection. To ensure the realism of the decision-making process, groups were required to actually execute their planned activity and verify with photographic evidence and transaction records. Ground truth labels were established by domain experts in travel behavior research, following factor definitions derived from utility-based choice modeling. **Figure 2** summarizes the x-GDP experiment process. For more detailed information regarding the experiment design and data characteristics, please refer to Parady et al. (2025b).

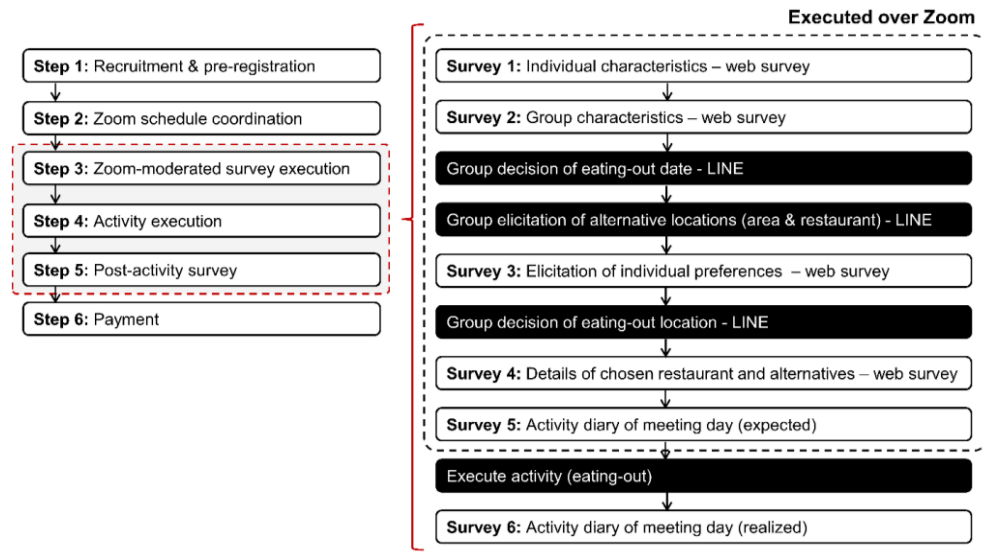

**Figure 2 Overview of x-GDP experiment process (source: Parady et al. (2025b)).**

### 3.2. Four-Step Prompting Framework

This study proposes a theory-driven, four-step prompting framework designed to convert unstructured group chat data into structured tabular data containing key decision-making factors. The framework sequentially executes *Entity Discovery* (Step 1), *Coreference Resolution* (Step 2), and *Relation Extraction* (Steps 3–4), with each step building on the outputs of its predecessors. Specifically, *Relation Extraction* is implemented in two stages, where Step 3 extracts preference polarity and Step 4 extracts the underlying attribute-based rationales. This sequential structure is designed to extract key factors required for utility-based models (i.e., decision makers, alternatives, and alternative-specific attributes). Furthermore, our method aims to capture latent individual traits (e.g., egocentrism) and implicit preferences towards alternatives, elements typically unobservable in traditional surveys, thereby enriching the modeling of joint behavior in activity-based models.

**Figure 3** presents an overview of the four-step framework and its associated prompting strategies. LLM performance at each step was evaluated using ground truth annotations and F1 scores. The optimal prompting technique was selected based on these metrics. **Figure 3** illustrates an example of the resulting step-specific selections, where CoT is selected for Step 1 and SR, PD, and MoRE are selected for Steps 2, 3, and 4, respectively, reflecting the increasing complexity of the reasoning tasks.

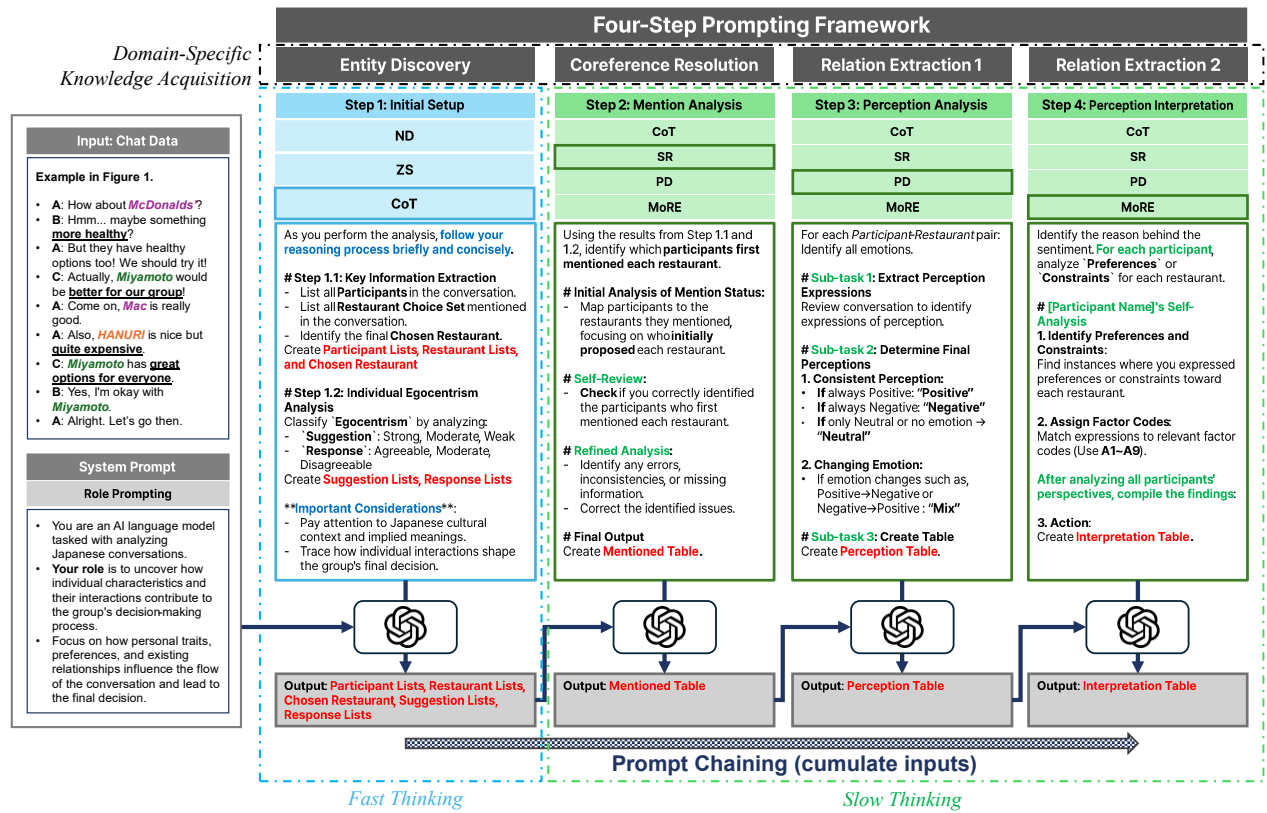

**Figure 3 Overview of the four-step framework and associated prompt engineering techniques, summarizing the prompts applied at each step and the resulting structured outputs based on the knowledge acquisition process.**

The four steps convert unstructured group chat data into structured tabular data containing both static and dynamic decision-making factors. Step 1 extracts static factors, including participants, alternatives, and egocentrism, while Steps 2–4 extract dynamic factors, such as choice set formation, evolving preferences, and their underlying rationales.

*Step 1.1: (Entity Discovery) Key Information Extraction*
This sub-step extracts key entities, including participants, restaurant alternatives, and the final decision, which serve as the foundation for subsequent analysis. The output is include:
- **Participant Lists**: All individuals involved in the group conversation.
- **Restaurant Lists**: All restaurant alternatives mentioned, suggested, or discussed.
- **Chosen Restaurant**: The final selection from the restaurant choice set.

*Step 1.2: (Entity Discovery) Individual Egocentrism Analysis*
This sub-step identifies each participant's decision-making style based on their degree of egocentrism. Egocentrism is assessed along two behavioral dimensions, where *Suggestion* means statements of preference and *Response* means statements in response to other people's preference:
- **Suggestion Lists**: Participants' suggestion strength, categorized as strong, moderate, or weak.
- **Response Lists**: The level of agreement or resistance toward others' suggestion, categorized as agreeable, moderate, or disagreeable.

*Step 2: (Coreference Resolution) Mention Analysis*
This step identifies the formation of the choice set throughout the conversation. While traditional choice models assume a fixed set of alternatives, group decision-making contexts reveal that alternatives often emerge dynamically. By resolving varied references to the same entity (e.g., *MacDonalds'* and *Mac* in **Figure 1**), Step 2 identifies which participant initially proposed each restaurant, revealing how participants shape the group's choice set. The output is:
- **Mentioned Table**: Indicates which participant first introduced each restaurant, categorized as mentioned or none.

*Step 3: (Relation Extraction) Perception Analysis*
This step extracts the relationship between each participant and restaurant alternatives by capturing emotional perceptions expressed in the chat. These perceptions offer a preliminary interpretative layer before the factor-based analysis in Step 4. Step 3 identifies the initial and final emotional responses of each participant, revealing how preferences evolve through social interaction. The output is:
- **Perception Table**: Records each participant's final perception of each restaurant, tracking whether sentiments remained consistent or shifted (categorized as positive, negative, neutral, or mix).

*Step 4: (Relation Extraction) Perception Interpretation*
This step extends the perception analysis by identifying the underlying reasons behind the perceptions identified in Step 3. While the previous step captured emotional tone, this step identifies the specific factors expressed explicitly or implicitly during the negotiation. Each participant's reasons are categorized into seven types of alternative-specific factors (A1-A7):
- **Restaurant Quality (A1)**: Includes food quality, food genre, service quality, restaurant ambiance, seating capacity, and the overall impression of the restaurant.
- **Accessibility and Location (A2)**: Covers ease of access, and the attractiveness of the surrounding area.
- **Schedule constraints (A3)**: Includes group members' schedules, business hours, and reservation availability.
- **Social utility for consensus (A4)**: Captures the extent to which a restaurant facilitates group agreement—not necessarily due to strong individual preference, but because it is broadly acceptable. This attribute reflects the restaurant's function as a social compromise that minimizes conflict.
- **Inertia (A5)**: Encompasses prior experience, familiarity, or variety-seeking attitudes.
- **Economic considerations (A6)**: Captures price range, individual/group budget constraints, and cost-effectiveness.
- **Others (A7)**: Includes any unclassified factors or cases lacking clear justification.

### 3.3. Prompt Engineering Techniques

Our framework incorporates two fundamental prompt engineering strategies. First, we apply *Role Prompting* by configuring the system prompt, the initial instruction defining the model's persona, to designate the LLM as a "Japan Conversation Analyst." This role-based setup provides essential domain-specific context, aligning the model's interpretation with the cultural norms and informal social interactions unique to Japan. Second, we implement *Prompt Chaining,* where outputs from earlier steps are cumulatively used as inputs for subsequent steps as shown in **Figure 3**. This approach supports cumulative reasoning and is designed to mitigate error propagation as the analysis advances from simple entity identification to complex contextual interpretation.

As shown in **Figure 4**, Step 1 employs two simple prompts (ND and ZS) and one reasoning prompt (CoT) to evaluate performance on *fast thinking* tasks. In contrast, Steps 2–4, which require higher-order cognitive processing, apply four reasoning prompts (CoT, SR, PD, MoRE) to identify the optimal method for structuring *slow thinking*. The complete prompt is provided in **Appendix B**. The effectiveness of each technique is evaluated quantitatively, as discussed in the following sections.

| System Prompt |
|---|
| ➤ **Role Prompting**<br>• Assigns the model as a "Japan Conversation Analyst"<br>• Highlights cultural context and informal language to understand |

| Step 1 |
|---|
| ➤ **No-Delimiters (ND) Prompting**<br>• Natural language to structure the task, without special characters or delimiters to break up the content. |
| ➤ **Zero-Shot (ZS) Prompting**<br>• Direct instructions without providing any examples, asking the model to extract key information based on pre-trained knowledge. |
| ➤ **Chain-of-Thought (COT) Prompting**<br>• Guides the model step-by-step through reasoning processes, prompting it to explain decisions briefly and concisely at each stage. |

| Step 2, 3, 4 |
|---|
| ➤ **Chain-of-Thought (CoT) Prompting (Wei et al., 2022)**<br>• Guides the model step-by-step through reasoning processes, prompting it to explain decisions briefly and concisely at each stage. |
| ➤ **Self-Refinement (SR) Prompting (Madaan et al., 2024)**<br>• Encourages the model to perform an initial analysis, review it, and refine the results through iterative refinement. |
| ➤ **Prompt Decomposition (PD) Prompting (Khot et al., 2022)**<br>• Breaks down complex tasks into simpler subtasks, each addressed through specialized prompts. |
| ➤ **Mixture of Reasoning Experts (MoRE) Prompting (Si et al., 2023)**<br>• Leverages multiple specialized reasoning models, each focusing on a specific domain, to collaboratively solve complex problems. |

**Figure 4** Overview of prompt engineering techniques at each step: ND, ZS, and CoT for Step 1, and CoT, SR, PD, and MoRE for Steps 2-4.

### 3.4. Performance Evaluation

The evaluation procedure systematically assesses the capability of LLM to interpret unstructured group chat data by comparing its outputs against the ground truth established by human annotators. The primary metric is the F1 score, balancing precision and recall to account for both omissions (i.e., false negatives) and hallucinated or extraneous outputs (i.e., false positives). For objective tasks with definitive answers (Steps 1.1 and 2), the F1 score directly reflects the model's accuracy. However, for inherently subjective tasks (Steps 1.2, 3, and 4), the F1 score measures the alignment between LLM interpretations and human judgment.

We define sets using standard notation, where $|S|$ is the cardinality of set $S$, $S_1 \cap S_2$ represents the intersection of sets $S_1$ and $S_2$. The ground truth data consists of a participant set $P^* = \{p_1, p_2, \cdots, p_n\}$ and a restaurant set $R^* = \{r_1, r_2, \cdots, r_m\}$. The F1 score is defined:

$$Precision = \frac{|S^{pred} \cap S^{true}|}{|S^{pred}|}, Recall = \frac{|S^{pred} \cap S^{true}|}{|S^{true}|}, F1 = 2 \times \frac{Precision \times Recall}{Precision + Recall} \quad (1)$$

Here, $S^{pred}$ represents the set of elements predicted by the LLM, while $S^{true}$ denotes the corresponding ground truth set. **Table 1** formalizes the structure of $S$ used in each step, specifying the basic unit and the domain of attribute values.

**Table 1 Output Format, Evaluation Unit, and Attribute Domains for Each Step**

| | Output | Basic Unit | The form of set $S$ | Attribute Value Sets |
|---|---|---|---|---|
| **Step 1.1** | Participant Lists | Element | $P = [p_1 \cdots p_n]^T_{n \times 1}$, where $S^{true} = P^*$ | $p_i \in \{Participants\ of\ Group\}$ |
| | Restaurant Lists | Element | $R = [r_1 \cdots r_m]^T_{m \times 1}$ where $S^{true} = R^*$ | $r_j \in \{Restaurant\ options\}$ |
| | Chosen Restaurant | Element | $r_k$ | $r_k = (Final\ Choice)$ |
| **Step 1.2** | Suggestion Lists | Pair | $\begin{bmatrix}(p_1, suggestion_1)\\ \vdots \\ (p_n, suggestion_n)\end{bmatrix}_{n \times 1}$ | $suggestion_i \in \{Strong, Moderate, Weak\}$ |
| | Response Lists | Pair | $\begin{bmatrix}(p_1, response_1)\\ \vdots \\ (p_n, response_n)\end{bmatrix}_{n \times 1}$ | $response_i \in \{Agreeable, Moderate, Disagreeable\}$ |
| **Step 2** | Mentioned Table | Triplet | $\begin{bmatrix}(p_1, r_1, mention_{11}) & \cdots & (p_1, r_m, mention_{1m})\\ \vdots & \ddots & \vdots \\ (p_n, r_1, mention_{n1}) & \cdots & (p_n, r_m, mention_{nm})\end{bmatrix}_{n \times m}$ | $mention_{ij} \in \{Mentioned, None\}$ |
| **Step 3** | Perception Table | Triplet | $\begin{bmatrix}(p_1, r_1, perception_{11}) & \cdots & (p_1, r_m, perception_{1m})\\ \vdots & \ddots & \vdots \\ (p_n, r_1, perception_{n1}) & \cdots & (p_n, r_m, perception_{nm})\end{bmatrix}_{n \times m}$ | $perception_{ij} \in \{Positive, Negative, Neutral, Mix\}$ |
| **Step 4** | Interpretation Table | Multi-label Triplet | $\begin{bmatrix}(p_1, r_1, factors_{11}) & \cdots & (p_1, r_m, factors_{1m})\\ \vdots & \ddots & \vdots \\ (p_n, r_1, factors_{n1}) & \cdots & (p_n, r_m, factors_{nm})\end{bmatrix}_{n \times m}$ | $factors_{ij} \subset \{A1, \cdots, A7\}$ |

Step 1 assesses whether the model correctly identifies participants, restaurant options, and individual-level egocentric behaviors:

$$Score_{Step1.1} = \frac{F1_{Participant\ Lists} + F1_{Restaurnat\ Lists} + F1_{Chosen\ Restaurant}}{3} \quad (2)$$

$$Score_{Step1.2} = \frac{F1_{Suggestion\ Lists} + F1_{Response\ Lists}}{2} \quad (3)$$

where $F1_{Participant\ Lists}$ evaluates the model's generated $P$ against ground truth $P^*$, $F1_{Restaurant\ Lists}$ evaluates the generated $R$ against ground truth $R^*$, $F1_{Chosen\ Restaurant}$ provides exact match scoring for the final group decision, $F1_{Suggestion\ Lists}$ and $F1_{Response\ Lists}$ evaluate generated set of (participant, egocentrism) pairs, where egocentrism is divided into two variables: *suggestion* and *response*, as shown in **Table 1**.

Steps 2 through 4 expand the evaluation framework to a three-dimensional level, assessing the model's ability to understand structured relationships between participants and restaurant alternatives. Each evaluation step operates on a triplet structure $(p_i, r_j, \alpha_{ij})$, where $\alpha_{ij}$ refer to either $mention_{ij}$, $perception_{ij}$, or $factors_{ij}$, depending on the analysis step:

$$Score_{Step\ k} = \begin{cases} F1_{Mention\ Table}, & for\ Step\ 2 \\ F1_{Perception\ Table}, & for\ Step\ 3 \\ Positive - F1_{Interpretation\ Table}, & for\ Step\ 4 \end{cases}, \quad where\ k = 2,3,4 \quad (4)$$

Unlike Steps 2 and 3, Step 4 requires the model to identify multiple factors for each (participant, restaurant) pair. Furthermore, since many $factors_{ij}$ in $S^{pred}$ and $S^{true}$ may correspond to cases where no factor was actually mentioned in the conversation (i.e., true label = *None*), using the standard $F1$ may overestimate LLM's performance. To address these, we adopt the $Positive - F1$, which calculates the average $F1$ only for entries where at least one factor exists in the ground truth's $factors_{ij}$. Let $\mathcal{D}_{positive}$ denotes the set of all $(p_i, r_j, factors_{ij})$ triplets where the ground truth is not empty:

$$Positive - F1 = \frac{1}{|\mathcal{D}_{positive}|} \sum_{i \in \mathcal{D}_{positive}} F1_i, \quad \text{where } \mathcal{D}_{positive} = \{i \mid |S_i^{true}| > 0\} \quad (5)$$

Each $F1_i$ is computed as:

$$Precision_i = \frac{|S_i^{pred} \cap S_i^{true}|}{|S_i^{pred}|}, \quad Recall_i = \frac{|S_i^{pred} \cap S_i^{true}|}{|S_i^{true}|}, \quad F1_i = 2 \times \frac{Precision_i \times Recall_i}{Precision_i + Recall_i} \quad (6)$$

Here, $Precision_i$ and $Recall_i$ represent how many of the predicted factors were correct, and how many of the correct factors were found, for each (participant, restaurant) pair $i$.

To ensure a rigorous evaluation of each step and reduce the impact of cumulative errors, we adopt a controlled selection strategy based on multiple iterations. We first identify the prompting technique with the highest average F1 score across its five runs. From this method, the single best-performing iteration is selected at each step (Equations 2 to 4), and its output serve as the input for the next step in the prompt chaining process. This design ensures each step's reasoning performance to be evaluated with minimal bias from upstream error propagation.

## 4. Results

This section presents both quantitative and qualitative evaluations of the LLM's (GPT-5) performance across the four steps of the proposed framework, each guided by different prompt engineering strategies. We first present quantitative results based on F1 scores, followed by a qualitative error analysis to identify and interpret key limitations of the model. For a detailed comparison between GPT-5 and GPT-4o, refer to **Appendix A**.

### 4.1. Quantitative Analysis

We conducted a quantitative analysis of GPT-5's performance across the four-step framework. **Table 2** and **Table 3** present the resulting F1 score (and Positive-F1 for Step 4).

**Table 2 Resulting F1 Scores for Step 1**

| Step 1 | | GPT-5 | | | |
|---|---|---|---|---|---|
| | | ND | ZS | CoT | Ave. |
| Participant Lists | Ave | 1.00 | 1.00 | 1.00 | 1.00 |
| | (Std.) | (0.00) | (0.00) | (0.00) | (0.00) |
| Restaurant Lists | Ave | 1.00 | 1.00 | 1.00 | 1.00 |
| | (Std.) | (0.02) | (0.01) | (0.02) | (0.02) |
| Chosen Restaurant | Ave | 0.92 | 0.96 | 0.95 | 0.95 |
| | (Std.) | (0.23) | (0.12) | (0.19) | (0.18) |
| Suggestion Lists | Ave | 0.69 | 0.67 | 0.65 | 0.67 |
| | (Std.) | (0.23) | (0.23) | (0.23) | (0.23) |
| Response Lists | Ave | 0.38 | 0.42 | 0.44 | 0.41 |
| | (Std.) | (0.31) | (0.29) | (0.26) | (0.28) |

**Table 3 Resulting F1 Scores for Steps 2-3 and Positive-F1 Scores for Step 4**

| Steps 2-4 | | GPT-5 | | | | |
|---|---|---|---|---|---|---|
| | | CoT | SR | PD | MoRE | Ave. |
| Mentioned Table (Step 2) | Ave. | 0.99 | 0.99 | 0.99 | 0.99 | 0.99 |
| | (Std.) | (0.03) | (0.03) | (0.02) | (0.03) | (0.02) |
| **Perception Table (Step 3)** | Ave. | 0.83 | 0.82 | 0.82 | 0.82 | 0.82 |
| | (Std.) | (0.14) | (0.15) | (0.14) | (0.15) | (0.14) |
| **Interpretation Table (Step 4)** | Ave. | 0.37 | 0.40 | 0.38 | 0.39 | 0.38 |
| | (Std.) | (0.21) | (0.20) | (0.21) | (0.20) | (0.20) |

Note: Ave. means Average of F1 score, Std. means Standard deviation of F1 score.

For tasks with definitive answers, the LLM demonstrates high accuracy overall. The model identified all *Participant Lists* and *Restaurant Lists* (both F1=1.00). However, performance declines slightly in the most crucial task of identifying the group's final decision, the *Chosen Restaurant*, where prompting techniques show some variation with ZS and CoT (F1 = 0.95–0.96) outperforming ND (F1 = 0.92). Since identifying the final choice requires tracing the flow of conversation to understand how the group reached consensus, prompts using structural delimiters may facilitate a more effective tracking of the dialogue than unstructured plain prompts.

In Step 2, the LLM maintains high levels of correctness (F1=0.99) with minimal variation across prompting strategies. This task requires linking varied references to the same restaurant. Once coreferences are resolved, detecting whether a participant initially mentioned a restaurant becomes a straightforward extraction task, aligning well with the model's proficiency in surface-level information retrieval.

In contrast, the LLM shows substantial difference from human annotation in subjective tasks. In Step 1.2, *Suggestion Lists* yield higher consistency than *Response Lists* (F1 = 0.67 vs. 0.41 for average), indicating a significant gap between the model's and humans' interpretations of egocentrism. Egocentrism in suggestions is often expressed explicitly, making it easier for the model to detect, whereas egocentrism in responses tends to be more context-dependent and implicit, which requires a deeper understanding of how individuals react to others in conversation. Consequently, the low F1 scores compared to explicit tasks, most notably in *Response Lists*, provide evidence of the model's difficulty in handling nuanced aspects of group dialogue that go beyond surface-level textual cues.

In Step 3, the model shows partial alignment with human annotations when evaluating each participant's perception toward restaurants (F1 = 0.82–0.83). This range of F1 scores may arise from the nature of the task itself. Unlike typical sentiment analysis that focuses on individual expressions, this task involves perceptions shaped through social interaction, where multiple participants influence one another. Such complexity may lead to interpretive differences between the model and humans, as perceptions are often shaped indirectly through group dynamics rather than through explicitly emotive language.

Step 4 presents the greatest challenge. The model is tasked with inferring the underlying reasons behind each perception, based on indirect cues embedded in the conversation. Across all prompting strategies, performance remains low (Positive-F1 = 0.37–0.40), indicating that the model infrequently selects the same reasoning factors as humans. These low scores suggest that extracting latent intent from unstructured group dialogue may go beyond what the current LLM is able to interpret.

In summary from the perspective of knowledge acquisition, *Entity discovery* (Step 1) shows high alignment with human annotation when extracting information directly stated in text, such as participants, alternatives, and final decisions, but alignment decreases when interpreting behavioral patterns such as egocentrism. *Coreference resolution* (Step 2) shows consistently high alignment. *Relation extraction* (Steps 3–4) shows the widest variation: interpreting perceptions yields moderate alignment, but identifying the underlying reasons shows substantial divergence from human judgment. These results suggest that the gap between LLM and human annotation widens progressively as the task shifts from surface-level extraction to the interpretation of expressed content, and finally to inference of unstated intent.

This quantitative analysis indicates that while the LLM has high capability for extracting definitive information, its ability to analyze the implicit and complex context of group interactions remains limited, as its interpretation often fails to align with that of humans.

### 4.2. Qualitative Analysis

We analyzed the model's output for each group by direct comparison with the original chat logs, especially focusing on groups with low accuracy. This analysis aims to pinpoint specific contextual situations where the LLM underperforms.

In the *Restaurant Lists* of Step 1.1, only 1.0% of all restaurants (3 out of 308) were not consistently identified across all iterations (five runs); however, even for these cases, the identification accuracy remained above 0.50, meaning they were still captured in the majority of runs. This result indicates that the LLM can easily identify terms that explicitly appear in the chat data.

In the *Chosen Restaurant* of Step 1.1, the LLM failed to consistently identify the final choice in 17.0% of the groups (8 out of 47) across all iterations. These misidentifications stem from the varied ways in which chosen restaurants were referenced. Among total sample, 21 restaurants were explicitly mentioned by their names or URL, 24 were referred to by their genre (e.g., "Italian", as shown in **Figure 5**), one was referenced by the proposer's name, and one was identified by its location. The aggregated error results across iterations are summarized below.

For restaurants explicitly mentioned by their names or URL, the error rate was 6.7%. Notably, 90.5% of these errors occurred in two groups with disrupted conversational flows. In the first group, a separate discussion about a different restaurant was inserted when the final decision was nearly made. In the second group, a URL for a different restaurant was provided post-decision because participants realized they had forgotten to include the required link earlier.

For restaurants mentioned by their genre, the error rate was 4.4%. Specifically, 31.3% of these errors were mislabeled as "*Not specified*". While the model often inferred the correct restaurant from genre alone, it occasionally defaulted to "*Not specified*". Furthermore, as illustrated in **Figure 5**, errors increased when the choice set contained multiple restaurants of the same genre (e.g., two Italian options). This suggests that the LLM may struggle to distinguish the intended alternative when explicit identifiers are absent and multiple options share similar attributes. Correct identification in these cases would have required the LLM to synthesize contextual cues, such as proximity to a specific location.

For the restaurant mentioned by the name of the proposer, the error rate was 6.7%, while the rate for location-based references was 0.0%. This indicates that the model is generally capable of associating alternatives with their proposers or spatial contexts.

In conclusion, while the absolute number of misidentifications is relatively small, this case analysis highlights a critical limitation: the LLM's difficulty in resolving coreferences when explicit identifiers (i.e., restaurant names) are absent, conversational flow is complex, or overlapping options exist in the choice set. In such situations, human-like contextual reasoning remains essential for accurate identification.

| Chat | Restaurant Lists | Chosen Restaurant |
|---|---|---|
| **A**: I prefer **Korean** & **Italian**. What do you think?<br>**B**: I ate **Korean** last week, so I would like to go to **Italian**.<br>**A**: It's ok for me. How about **C & D**?<br>**C**: **Italian** is **not so far** from our college, so I think it's good.<br>**D**: **Me too!** Let's go to **Italian**! | 1. Hanuri (**Korean**)<br>2. **Napoli Pizza (Italian)** ←<br>3. MacDonald<br>4. **Kamakura Pasta (Italian)** ←<br>5. Edomae Zushi | **True Answer**<br>≠<br>**LLM's output** |

**Figure 5 Chat whose choice set includes the same genre of the chosen restaurant**

In Step 1.2, while *Suggestion Analysis* achieved 0.67 of F1 score, *Response Analysis* exhibited significantly lower performance, 0.41 (**Figure 6**). To investigate this discrepancy, we compared the human annotations with LLM outputs for both tasks. For *Suggestion Analysis*, an examination of the five lowest-performing groups revealed that 45.9% of the suggestions labeled as *Weak* in the human annotations were classified as *Moderate* by the LLM (**Figure 7a**). This indicates that the LLM tends to overestimate the assertiveness of participants' suggestions. Similarly, in Response Analysis, 96.0% of the responses labeled "Moderate" by humans were classified as "Agreeable" by the LLM (**Figure 7b**). Taken together, these findings suggest that the LLM systematically perceives individuals in group discussions as more cooperative than they actually are, overestimating both the proactiveness of suggestions and the positivity of responses (Strachan et al., 2024).

In Step 2, the average F1 score was 0.98. 5.2% of all restaurants (16 out of 308) were not consistently associated with their correct proposers across all iterations. Among these, three exhibited identification accuracies below 0.5, including one with an accuracy of 0. Conversely, seven restaurants achieved an accuracy of 0.95, indicating an incorrect output in only a single iteration.

For seven restaurants, the LLM mistakenly omitted or incorrectly mapped names in the Mentioned Table. This suggests that the LLM occasionally encounters failures during the Prompt Chaining process, where it fails to correctly propagate the *Restaurant Lists* from Step 1.1 to subsequent steps.

A detailed review of the chat for the restaurant with an accuracy of 0 reveals that it was mentioned exclusively via an URL, whereas other restaurants were referenced by both names and URL. Furthermore, the restaurant's name was only introduced much later in the discussion by a different participant. This suggests that the LLM struggles with inconsistent referencing styles and temporal distance between an URL and its corresponding name, rather than failing to recognize the restaurant itself.

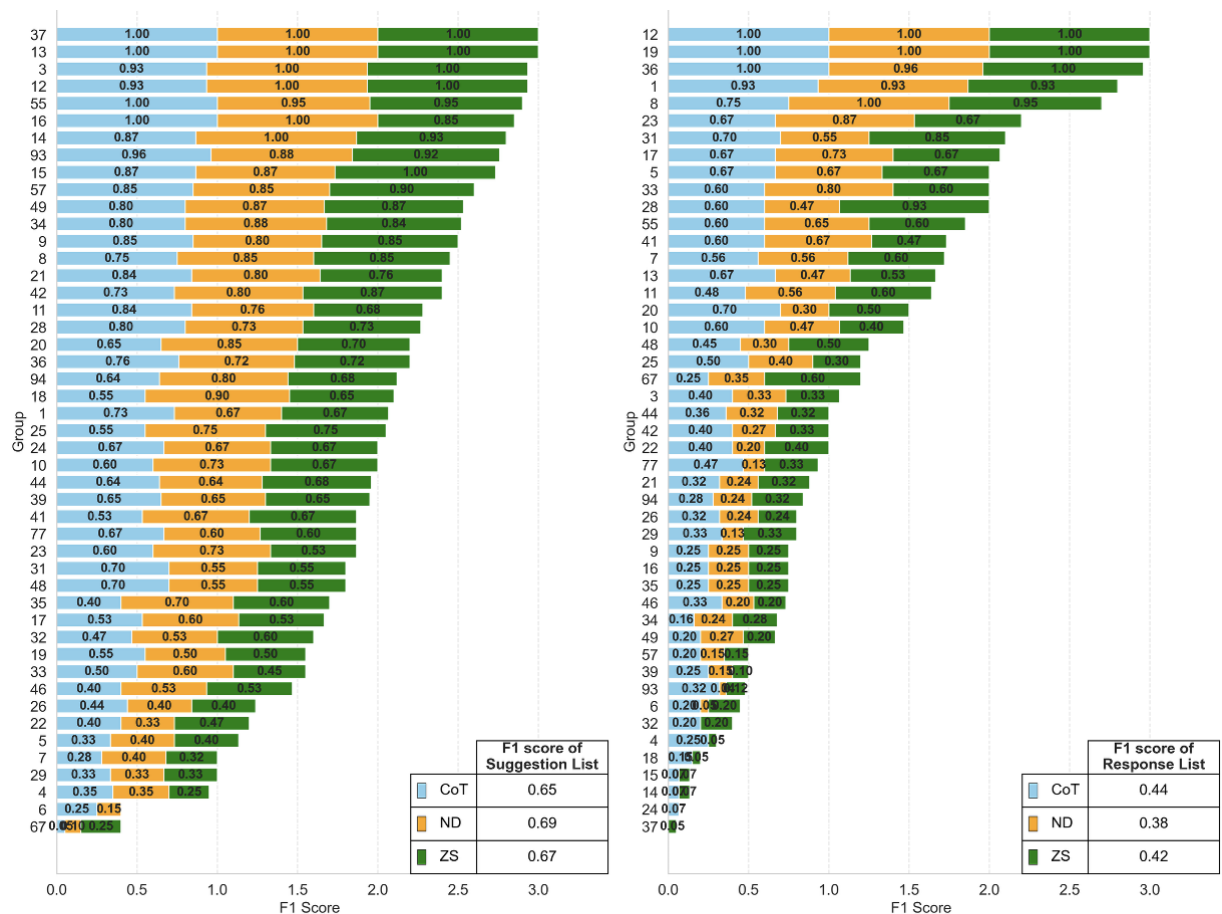

**Figure 6** Score bar of *Suggestion Analysis* **(left)** and *Response Analysis* **(right)**

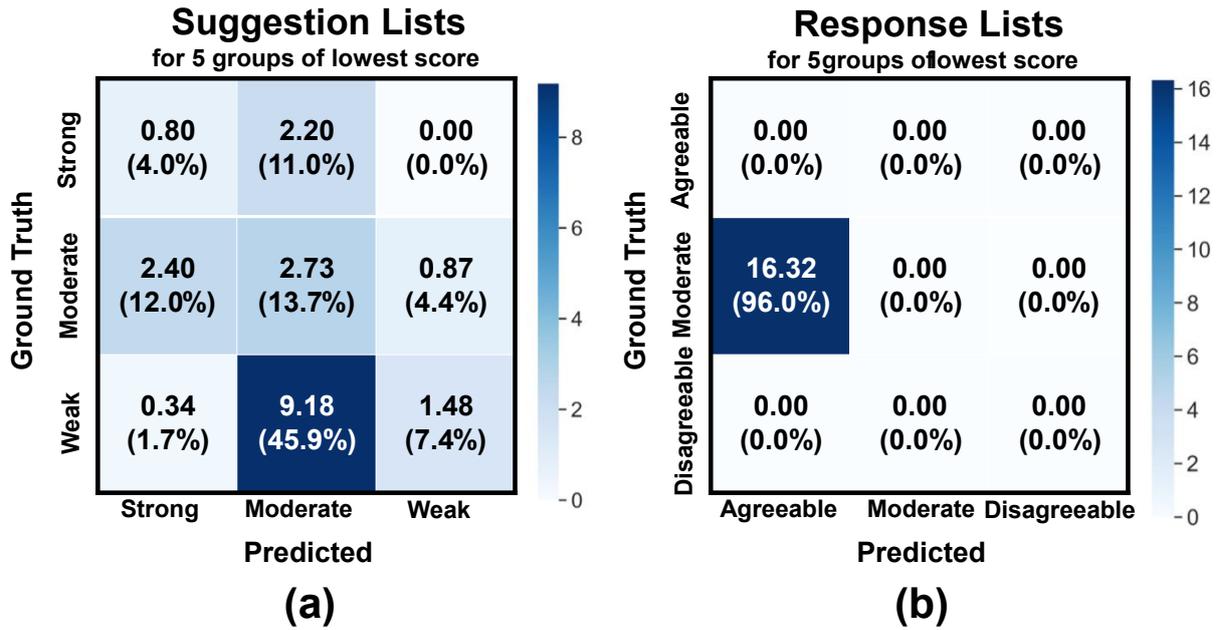

**Figure 7 Ground truth and LLM's predicted output matrix of** *Suggestion Analysis* **(left) and** *Response Analysis* **(right)**

In Step 3, the average F1 score was 0.73. Across the entire dataset, the primary source of error was the misclassification between *Positive* and *Neutral* perceptions: 3.5% of instances labeled as *Positive* in the ground truth were classified as *Neutral* by the model, while 9.7% of *Neutral* instances were misclassified as *Positive* (**Figure 8**). This tendency was consistently observed across groups.

However, a more pronounced bias emerged when focusing on the five lowest-scoring groups: 13.5% of *Neutral* instances were misclassified as *Positive*, and 8.3% of *Negative* instances were misclassified as *Neutral*. These results further demonstrate that LLMs tend to interpret human perception expressions toward restaurants more positively than human annotators intended. This finding is consistent with the results observed in Step 1.2, reinforcing the conclusion that LLMs possess a general positivity bias when interpreting unstructured social dialogues.

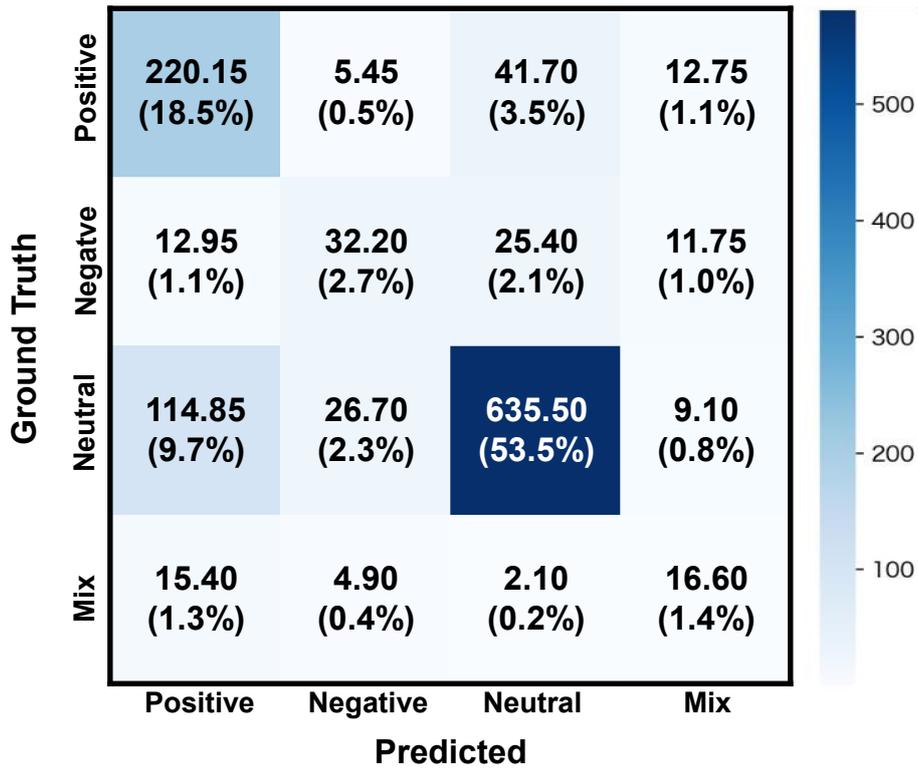

**Figure 8** Ground truth and LLM's predicted output matrix of *Perception Analysis*

      In Step 4, the average F1 score was approximately 0.44. Examining performance by factor, the LLM correctly predicted 48.4% for A1 (Restaurant Quality), 48.6% for A2 (Accessibility), 70.8% for A3 (Schedule Constraints), 36.7% for A4 (Social utility for consensus), 52.4% for A5 (Inertia), 54.7% for A6 (Economic Constraint) and 3.6% for A7 (Others) (**Figure 9**). Overall, the model struggled to identify the reasons behind participants' perceptions, confirming that this task is particularly challenging for LLMs. Performance varied significantly across factors: A1, A2, A3, A5 and A6, which tend to be explicitly mentioned in chats, were predicted with higher accuracy. In contrast, A4 showed low accuracy, likely because it requires a nuanced understanding of interpersonal dynamics that are difficult to infer from text alone. For A7, many instances were misclassified as A1, resulting in exceptionally low accuracy for this category. These results suggest a clear dichotomy between identifiable, explicit factors and ambiguous, implicit reasoning factors.

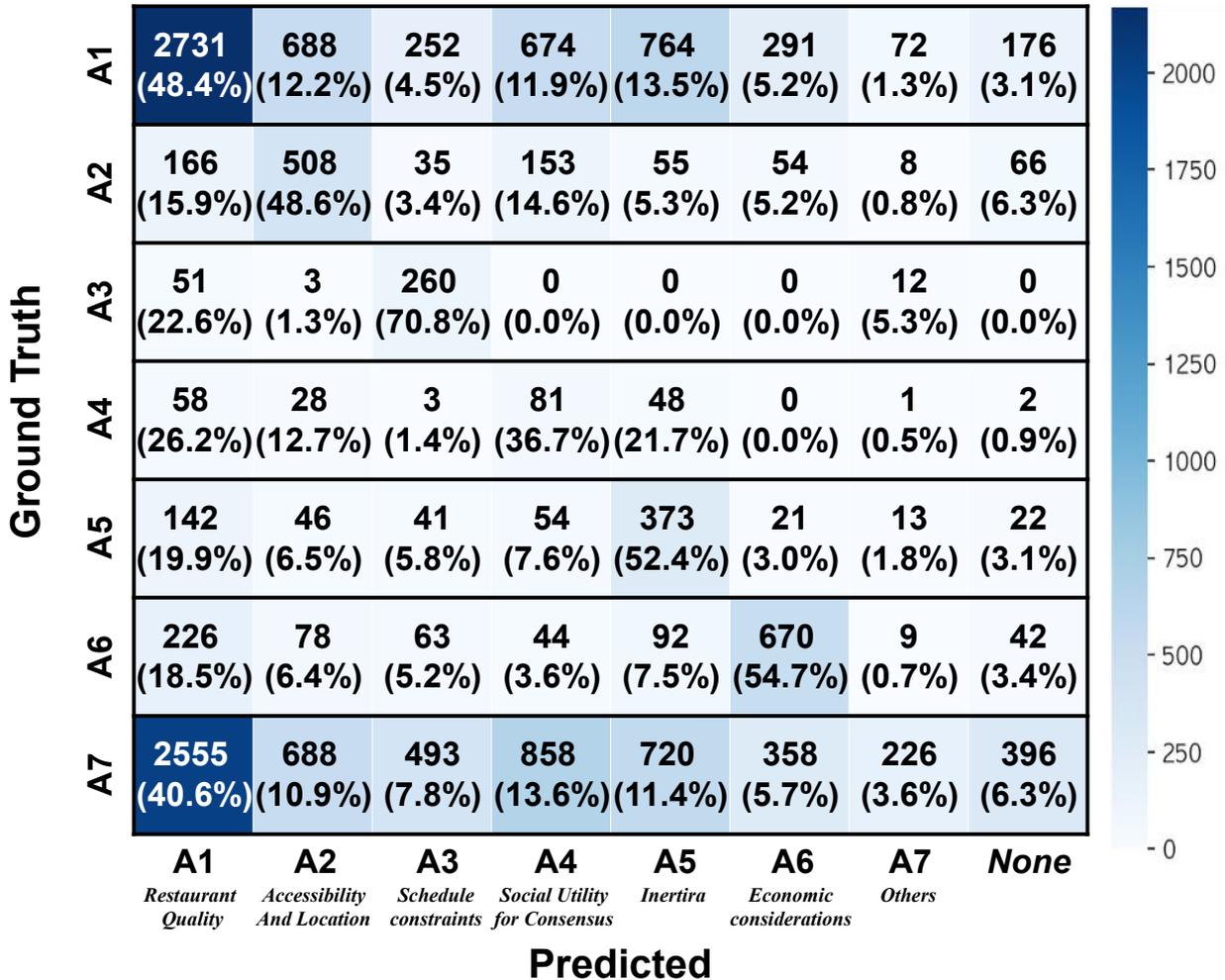

**Figure 9** Ground truth and LLM's predicted output matrix of *Perception Interpretation*

In summary, the qualitative analysis reveals three systematic patterns in LLM errors. First, the model exhibits an "optimism bias," consistently interpreting social interactions as more cooperative and positive than human annotators; it overestimates both the assertiveness of suggestions and the agreeableness of responses. Second, the model's performance diminishes when identifying the chosen restaurant if explicit identifiers (i.e., restaurant names) are absent, conversational flow is fragmented, or multiple alternatives share overlapping attributes. Third, the model shows a limited capacity for inferring underlying motivations, particularly for factors requiring an understanding of interplay between participants (A4: Social utility for consensus) or those lacking explicit justification (A7: Others). These patterns suggest that while LLMs can reliably extract surface-level information, their interpretative depth remains constrained for tasks involving socially grounded reasoning and implicit intent.

## 5. Conclusion

This study proposed and evaluated a structured four-step framework grounded in knowledge acquisition, the foundational stage of Knowledge Graph (KG) construction, to interpret group decision-making processes from unstructured chat data using a large language model (LLM). By decomposing the extraction process into *Entity Discovery*, *Coreference Resolution*, and *Relation Extraction*, the framework provides structured guidance that specifies the information the LLM must extract at each step. The framework progresses from surface-level information extraction to the interpretation of participants' preferences and their underlying determinants, enabling systematic assessment of the model's ability to process increasingly complex cognitive tasks within group conversations.

Quantitatively, the LLM demonstrated high performance in deterministic tasks such as identifying key entities (Step 1.1) and resolving coreferences (Step 2). However, performance declined significantly in tasks requiring the interpretation of nuanced social dynamics, such as recognizing egocentric behavior (Step 1.2), evaluating individual perception based on group interactions (Step 3), and inferring preference rationales (Step 4). Notably, prompting strategies had limited influence on performance, indicating that the primary challenge lies in the model's inherent ability to extract latent behavioral signals, rather than the specific instructions provided.

Qualitative analysis further revealed the limitations of LLMs in interpreting human intent under complex conditions. While the model reliably identified surface-level information, it struggled with the nuanced expression of social preferences and the inference of emotional perception. Furthermore, the LLM exhibited a systematic "optimism bias," interpreting interactions as more cooperative and emotionally positive than human annotators judged them to be. These tendencies suggest that providing unstructured chat data alone may be insufficient to capture the full depth of group decision-making dynamics.

Overall, these findings highlight fundamental limitations in current LLMs when applied to group decision-making processes. Despite the use of prompt chaining and structured guidance, model performance declines in tasks involving ambiguity, implicit meaning, or socially grounded reasoning. While LLMs are effective at extracting definitive elements from dialogue, their interpretive capacity remains limited for reliably modeling dynamic social interactions. Consequently, the use of LLMs to infer behavioral signals from group chat data should be approached with caution, especially when such outputs inform downstream activity-based models. Future research should not only aim to improve model accuracy, but also critically assess the boundaries of LLM applicability in socially nuanced contexts.

**Appendix A. Comparison between GPT-5 and GPT-4o**
**A1. Quantitative Analysis**
To examine whether advancements in LLM capabilities improve performance on implicit reasoning tasks, we compare GPT-5 and GPT-4o across the four-step framework. **Table A1** and **Table A2** present the resulting F1 scores.

**TABLE A1. Results for Step 1**

| Step 1 | | GPT-5 | | | | GPT-4o | | | |
|---|---|---|---|---|---|---|---|---|---|
| | | ND | ZS | CoT | Ave. | ND | ZS | CoT | Ave. |
| Participant Lists | Ave | 1.00 | 1.00 | 1.00 | **1.00** | 1.00 | 1.00 | 1.00 | **1.00** |
| | (Std.) | (0.00) | (0.00) | (0.00) | (0.00) | (0.00) | (0.00) | (0.00) | (0.00) |
| Restaurant Lists | Ave | 1.00 | 1.00 | 1.00 | **1.00** | 0.98 | 0.97 | 0.97 | 0.97 |
| | (Std.) | (0.02) | (0.01) | (0.02) | (0.02) | (0.07) | (0.10) | (0.10) | (0.09) |
| Chosen Restaurant | Ave | 0.92 | 0.96 | 0.95 | **0.95** | 0.90 | 0.90 | 0.94 | 0.91 |
| | (Std.) | (0.23) | (0.12) | (0.19) | (0.18) | (0.30) | (0.29) | (0.23) | (0.27) |
| Suggestion Lists | Ave | 0.69 | 0.67 | 0.65 | **0.67** | 0.66 | 0.70 | 0.68 | **0.67** |
| | (Std.) | (0.23) | (0.23) | (0.23) | (0.23) | (0.21) | (0.21) | (0.20) | (0.20) |
| Response Lists | Ave | 0.38 | 0.42 | 0.44 | **0.41** | 0.26 | 0.28 | 0.29 | 0.27 |
| | (Std.) | (0.31) | (0.29) | (0.26) | (0.28) | (0.34) | (0.31) | (0.30) | (0.32) |

**TABLE A2. Results for Steps 2 and 3, Positive-F1 Score for Step 4**

| Steps 2-4 | | GPT-5 | | | | | GPT-4o | | | | |
|---|---|---|---|---|---|---|---|---|---|---|---|
| | | CoT | SR | PD | MoRE | Ave. | CoT | SR | PD | MoRE | Ave. |
| Mentioned Table | Ave. | 0.99 | 0.99 | 0.99 | 0.99 | **0.99** | 0.92 | 0.55 | 0.91 | 0.96 | 0.83 |
| | (Std.) | (0.03) | (0.03) | (0.02) | (0.03) | (0.02) | (0.15) | (0.22) | (0.16) | (0.07) | (0.15) |
| Perception Table | Ave. | 0.83 | 0.82 | 0.82 | 0.82 | **0.82** | 0.76 | 0.64 | 0.74 | 0.77 | 0.73 |
| | (Std.) | (0.14) | (0.15) | (0.14) | (0.15) | (0.14) | (0.21) | (0.28) | (0.22) | (0.20) | (0.23) |
| Interpretation Table | Ave. | 0.37 | 0.40 | 0.38 | 0.39 | **0.38** | 0.37 | 0.37 | 0.37 | 0.37 | 0.37 |
| | (Std.) | (0.21) | (0.20) | (0.21) | (0.20) | (0.20) | (0.25) | (0.25) | (0.26) | (0.25) | (0.25) |

Note: Ave. means Average of F1 score, Std. means Standard deviation of F1 score. The bold indicates the higher average score between GPT-5 and GPT-4o.

In Step 1.1, GPT-5 shows improvement over GPT-4o in *Restaurant Lists* (F1=1.00 vs. 0.97) and *Chosen Restaurant* (F1=0.95 vs. 0.91), while *Participant Lists* remain perfect for both models. In Step 1.2, *Suggestion Lists* yield similar results (F1=0.67 for both), but *Response Lists* show higher score with GPT-5 (F1=0.41 vs. 0.27). Since responses to others' suggestions are expressed more indirectly than one's own suggestions, this improvement may reflect GPT-5's enhanced ability to understand contextual cues in conversation.

In Steps 2–4, GPT-5 demonstrates more consistent performance across prompting strategies. Most notably, GPT-4o shows high variance for *Mentioned Table* depending on prompting technique (F1=0.55–0.96), whereas GPT-5 maintains stable performance (F1=0.99) regardless of prompt design. *Perception Table* also improves with GPT-5 (F1=0.82 vs. 0.73). *Interpretation Table* shows minimal difference between the two models (Positive-F1=0.38 vs. 0.37), suggesting that inferring latent intent is not changed much regardless of model advancement.

These quantitative results indicate that GPT-5 shows higher alignment with human annotation and reduced sensitivity to prompt design compared to GPT-4o, while the gap between LLM and human annotation for implicit information persists across both models. To identify the specific sources of these differences, the following section conducts a qualitative analysis.

**A2. Qualitative Analysis**
The Qualitative Analysis aims to identify aspects in which GPT-4o demonstrates weaker performance compared to GPT-5.

In the *Restaurant Lists* of Step 1.1, 7.5% of all restaurants (23 out of 308) were misidentified by GPT-4o, 7.7 times higher than GPT-5. Two primary patterns of misidentification were observed.

First, GPT-4o tends to misidentify restaurants that are mentioned only via hyperlinks, compared to restaurants mentioned via combinations of restaurant names, addresses and hyperlinks. To investigate this, the chat data are categorized according to how alternatives were mentioned: single-type chats and mixed-type chats (**Figure A1**). Single-type chats refer to chats where all restaurant alternatives are mentioned in the same format (e.g., only "hyperlink" or only "name, address and hyperlink"). In contrast, mix-type chats include a combination of "hyperlink" and "name, address and hyperlink". Among the misidentified restaurants, 69.6% were mentioned via hyperlinks within mix-type chats. This suggests that GPT-4o struggles to identify alternatives when the formats how they are mentioned differ within the same chat. One possible reason is that GPT-4o tends to fixate on the most consistent reference style within a chat. Moreover, even when a list of hyperlinks and corresponding restaurant names is provided to the LLM for reference, LLMs still often fail to make the correct identification.

Second, GPT-4o fails to identify restaurants when multiple hyperlinks and comments are embedded in a single message. Specifically, four messages included two hyperlinks each, and one message included five hyperlinks along with detailed descriptions for each. While the eight restaurants mentioned in the four messages without any text were correctly identified, five restaurants in the single message were misidentified. This indicates that the combination of multiple alternatives and accompanying texts within the same message may confuse GPT-4o, making it more difficult to distinguish individual alternatives. In summary, these findings suggest that GPT-4o are more sensitive to the format and structure of how alternatives are presented in group chat data than GPT-5.

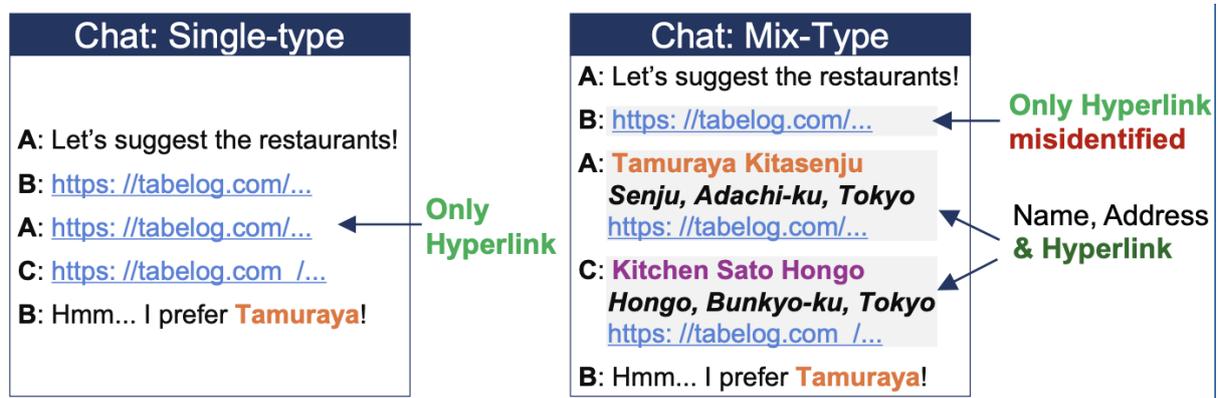

**Figure A1 Example of single-type chat and mix-type chat**

In the *Chosen Restaurant* of Step 1.1, the number of restaurants misidentified in at least one iteration by GPT-4o (5 out of 47) is lower than GPT-5 (8 out of 47). However, the total number of incorrect iterations produced by GPT-4o (61 out of 705) is higher than produced by GPT-5 (38 out of 705). When classified by mention type, the error rate for restaurants mentioned by their names or hyperlinks was 4.8%, which is slightly lower than that of GPT-5 (6.7%). In contrast, for restaurants mentioned by their genre, the error rate was 8.6%, which is approximately twice as high as that of GPT-5 (4.4%). Moreover, for restaurants mentioned by the name of the proposer, the error rate was 100%, whereas GPT-5 showed an error rate of 6.7%. These results suggest that GPT-4o tends to misidentify more frequently than GPT-5 when candidate names are not explicitly stated in the chat data.

In the *Suggestion Analysis* of Step 1.2, the F1 score of GPT-4o (0.68) was comparable to that of GPT-5 (0.67), however, the pattern of the errors differed between the two models. Suggestions labeled as *Moderate* in human annotations were often labeled as *Weak* by GPT-4o, which means GPT-4o tends to underestimate proactiveness of suggestions while GPT-5 tends to overestimate it. In *Response Analysis*, the F1 score of GPT-4o (0.28) was lower than that of GPT-5 (0.41), although the pattern of the error was similar across models. Responses labeled as *Moderate* in human annotations were often labeled as *Agreeable* by GPT-4o. This indicates that GPT-4o tends to overestimate the positivity of responses, which is the same trend of GPT-5.

In Step 2, 21.6% of all restaurants (66 out of 308) were not consistently associated with their proposers across all iterations, which is substantially higher than the corresponding rate for GPT-5 (5.2%). Among these restaurants, 26 exhibited accuracies below 0.5, including five restaurants with an accuracy of 0. For 25 restaurants, GPT-4o mistakenly included restaurant names in the Mention Table, which occurred more frequently than for GPT-5 (seven restaurants). Moreover, for eight restaurants, GPT-4o incorrectly identified two participants as proposers, a type of error that did not occur with GPT-5. These results suggest that GPT-4o is more prone to associating candidates with incorrect proposers and is less effective than GPT-5 at propagating outputs from earlier steps to later steps.

In Step 3, the main source of errors for GPT-4o was misclassification between *Positive* and *Neutral* perceptions: 8.5% of instances labeled as *Positive* in the ground truth were classified as *Neutral* by the model, and 7.0% of *Neutral* instances were misclassified as *Positive*. This pattern is similar to that observed for GPT-5. However, the results also show that GPT-4o more frequently misclassified instance labeled as *Positive* in the ground truth than GPT-5 (3.5%), and less frequently misclassified instances labeled as *Neutral* than GPT-5 (9.7%). This indicates that GPT-4o has a stronger tendency to underestimate positivity than GPT-5.

In Step 4, when performance was examined for each factor, 61.8% of instances were correctly predicted as A1 (Restaurant Quality), 37.0% as A2 (Accessibility), 60.0% as A3 (Schedule Constraints), 14.3% as A4 (Social Utility for Consensus), 48.2% as A5 (Inertia), 61.1% as A6 (Economic Constraint), and 0.57% as A7 (Others). Among A1 to A6, GPT-4o showed substantially lower accuracy for A4 than for the other labels. This tendency is similar to that observed for GPT-5, but the accuracy of GPT-4o is even lower. As for A7, many of the responses labeled as A7 in the ground truth were misclassified by the model as A1, even though participants did not mention restaurant quality. GPT-5 also misclassified A7 as A1, suggesting that both models have difficulties in classifying instances expressed using ambiguous expressions. These results indicate that GPT-4o struggles to infer underlying reasons for participants' sentiments, especially those that are not explicitly mentioned in the chat.

In summary, the comparison between GPT-5 and GPT-4o reveals that model advancement yields improvements in several aspects: GPT-5 demonstrates higher accuracy in explicit extraction tasks, greater robustness to input format variations, improved stability across prompting strategies, and more reliable propagation of outputs through prompt chaining. However, both models exhibit similar patterns of difficulty when interpreting implicit factors, particularly social consensus (A4) and factors beyond the predefined categories (A7). These findings suggest that while newer LLM versions enhance surface-level extraction capabilities, the fundamental challenge of inferring implicit social meaning from group conversations remains largely unresolved by model advancement alone.

## Appendix B. Full Prompts for Each Step and Prompting Technique

> **Prompt B.1 [Role Prompting] for** *System Prompt*
>
> You are an AI language model tasked with analyzing Japanese conversations where a group decides on a final food destination.
> Your role is to uncover how individual characteristics and their interactions contribute to the group's decision-making process.
>
> Focus on how personal traits, preferences, and existing relationships influence the flow of the conversation and lead to the final decision.
>
> **Important Considerations:**
> - Pay careful attention to the Japanese cultural context, including colloquialisms, abbreviations, and slang commonly found in informal conversations.
> - Be mindful that participants may have shared knowledge or assumptions not explicitly mentioned in the dialogue.
> - Interpret not just what is explicitly said, but also implied meanings based on shared history, tone, or familiarity between participants.
> - Maintain objectivity in your analysis while being sensitive to cultural nuances.
> - Your ultimate goal is to trace how individual interactions shape the group's final decision on the food destination.
> - In Conversation Text Data (Input), there are CONVERSATION PART and INFORMATION PART. INFORMARION PART includes `Website Link` and `Restaurant`. `Website Link` expresses website link appeared in CONVERSATION PART. `Restaurant` expresses the name of the restaurant corresponding to `Website Link`.
> - Some `Website Link` is blank because some restaurant names are expressed with no link.

**Prompt B.2 [No-Delimiters] for *Step 1***

The following is a conversation in Japanese between a group of friends trying to choose a restaurant to eat-out. Please analyse this conversation to identify how individual characteristics and interactions among group members affect the decision-making process.

**Step1.1.**
Focus on extracting all the **participants' names**, all the **restaurant names** mentioned during the conversation, and the name of the restaurant that is **finally chosen**. Be sure to extract restaurant names from the CONVERSATION PART, and extract the exact name referencing the INFORMATION PART.

**Step1.2.**
**First**, analyze egocentrism of each individual based on **the strength of each individual's suggestion**, where a suggestion refers to statements expressing a claim of preference. Egocentrism can be categorized as Strong, Moderate, or Weak. Strong egocentrism indicates that an individual frequently proposes their preference, often using strong expressions to insist on their choice. Moderate egocentrism suggests that a person does propose their preference, but not very often, and generally uses neutral expressions. Weak egocentrism refers to cases where an individual does not propose a preference or has no particular ideas.
**Second**, analyze egocentrism of each individual based on **the attribute of each individual's response**, which denotes statements given in reaction to others' preference claims. Egocentrism in responses can be classified as Agreeable, Moderate, or Disagreeable. An Agreeable response indicates that the individual often agrees with others' suggestions and is willing to change their opinion to align with them. Moderate responses show no strong tendency towards agreement or disagreement. Disagreeable responses are characterized by a frequent tendency to disagree with others' suggestions, with the individual showing resistance to changing their stance easily.

All outputs should be selected only from the provided options.

**Output for Step 1.1:** <Participant Lists>, <Restaurant Lists>, <Chosen Restaurant>
**Output for Step 1.2:** <Suggestion Lists>, <Response Lists>

**Prompt B.3 [Zero-Shot] for** *Step 1*

The following is a conversation in Japanese between a group of friends trying to choose a restaurant to eat-out. Please analyse this conversation to identify how individual characteristics and interactions among group members affect the decision-making process.

---
# Step 1.1: Initial Setup
Analyzing `Conversation` text data, Extract Key Information.

- `Participant`: List all participants' names.
- `Restaurant`: List all restaurant names mentioned. Extract `Restaurant` mentioned in the CONVERSATION PART and extract the "exact name" referencing the **INFORMATION PART**.
- `Final`: Identify and extract the chosen restaurant in the **INFORMATION PART**.

---
# Step 1.2: Individual Characteristics Analysis

## 1.2.1 Suggestion
Analyze suggestion's strongness. Suggestion means claims of preference.
- `Egocentrism` Types: Strong, Moderate, Weak
- `Strong`: Propose one's preference in **many** times. Insist on one's preference with **strong** expressions.
- `Moderate`: Propose one's preference, but **not many** times. Insist on one's preference with **normal** expressions.
- `Weak`: Don't propose one's preference, have **no ideas**.

## 1.2.2 Response
Analyze response's attribute. Response means statements in response to other people's claims of preference.
- `Egocentrism` Types: Agreeable, Moderate, Disagreeable
- `Agreeable`: **Often** agree with other's suggestion and change one's mind to follow the others.
- `Moderate`: **Not too much** tendency to agree and disagree.
- `Disagreeable`: **Often** disagree with other's suggestion and don't change one's mind easily.
---
**Note:**
- All outputs should be selected only from the provided options.
- Please adhere to the output format.

**Output for Step 1.1:** <Participant Lists>, <Restaurant Lists>, <Chosen Restaurant>
**Output for Step 1.2:** <Suggestion Lists>, <Response Lists>

**Prompt B.4. [Chain-of-Thought] for *Step 1***

The following is a conversation in Japanese between a group of friends trying to choose a restaurant to eat-out. Please analyse this conversation to identify how individual characteristics and interactions among group members affect the decision-making process.
As you perform the analysis, explain **your reasoning process** briefly and concisely in Japanese at each step.

---

# Step 1.1: Initial Setup
Extract Key Information.

- `Participant`: List all participants' names.
- `Restaurant`: List all restaurant names mentioned. Extract `Restaurant` mentioned in the CONVERSATION PART and extract the "exact name" referencing the **INFORMATION PART**.
- `Final`: Identify and extract the chosen restaurant in the **INFORMATION PART**.

---

# Step 1.2: Individual Characteristics Analysis

## 1.2.1 Suggestion
Analyze suggestion's strongness. Suggestion means claims of preference.
- `Egocentrism` Types: Strong, Moderate, Weak
- `Strong`: Propose one's preference in many times. Insist on one's preference with strong expressions.
- `Moderate`: Propose one's preference, but not many times. Insist on one's preference with normal expressions.
- `Weak`: Don't propose one's preference, have no ideas.

## 1.2.2 Response
Analyze response's attribute. Response means statements in response to other people's claims of preference.
- `Egocentrism` Types: Agreeable, Moderate, Disagreeable
- `Agreeable`: **Often** agree with other's suggestion and change one's mind to follow the others.
- `Moderate`: **Not too much** tendency to agree and disagree.
- `Disagreeable`: **Often** disagree with other's suggestion and don't change one's mind easily.

---

**Important Considerations:**
- Pay careful attention to the Japanese cultural context, including colloquialisms, abbreviations, and slang commonly found in informal conversations.
- Consider that participants may have shared knowledge or assumptions not explicitly mentioned in the dialogue.
- Interpret not just what is explicitly said, but also implied meanings based on shared history, tone, or familiarity between participants.
- Your ultimate goal is to identify how individual interactions shape the group's final decision.

**Note:**
- At each step, please explain your reasoning process briefly and concisely in Japanese.
- All outputs should be selected only from the provided options.
- Please adhere to the output format.

---

**Output for Step 1.1:** <Participant Lists>, <Restaurant Lists>, <Chosen Restaurant>
**Output for Step 1.2:** <Suggestion Lists>, <Response Lists>

**Prompt B.5 [Chain-of-Thought] for *Step 2***

# Step 2: Mention Analysis

Using the results from Steps 1.1 and 1.2, perform a detailed analysis to determine which participants first mentioned each restaurant.

**For each line of the conversation:**
  - Identify if a restaurant is mentioned.
  - Note which participant mentioned it.
  - Determine if this is the first time the restaurant is suggested.
  - Record any context or reasoning provided.

**Construct a chain of thought:**
  - Step-by-step, build upon the information gathered to map out who first mentioned each restaurant.
  - Include your reasoning at each step to show how you identified the initial mentions.

**Final Output Format:**
MentionedTable
- Rows: Participants extracted in Step 1.1.
- Columns: Restaurants extracted in Step 1.1.
- Fill each cell with "Mentioned" if the participant **first mentioned** the restaurant, or "None" if not.

---
**Note:**
- Generate a comprehensive chain of reasoning leading to the final MentionedTable.
- Ensure that each reasoning step is clear and logically follows from the previous steps.
- Focus on identifying who **first** mentioned each restaurant.
- Adhere strictly to the output format.

**Output for Step 2:** <Mentioned Table>

**Prompt B.6 [Self-Refinement] for *Step 2***

# Step 2: Mention Analysis

Using the results from Steps 1.1 and 1.2, identify which participants first mentioned each restaurant.

**Initial Analysis:**
  - Map participants to the restaurants they mentioned, focusing on who initially proposed each restaurant.

**Self-Review:**
  - Critically assess your initial analysis for accuracy and completeness.
  - Check if you correctly identified the participants who first mentioned each restaurant.
  - Identify any errors, inconsistencies, or missing information.

**Refined Analysis:**
  - Correct the identified issues.
  - Present the improved mapping, ensuring you highlight who made the initial mentions.

**Final Output Format:**
MentionedTable
- Rows: Participants extracted in Step 1.1.
- Columns: Restaurants extracted in Step 1.1.
- Fill each cell with "Mentioned" if the participant **first mentioned** the restaurant, or "None" if not.

---
**Note:**
- Clearly separate the initial analysis, self-review, and refined analysis.
- Document any changes made during refinement.
- Pay careful attention to identifying the participants who first mentioned each restaurant.
- Adhere strictly to the final output format.

**Output for Step 2:** <Mentioned Table>

**Prompt B.7 [Prompt Decomposition] for *Step 2***

# Step 2: Mention Analysis

Decompose the analysis into the following detailed steps to accurately determine which participants first mentioned each restaurant.

## Sub-task 1: Extract Initial Restaurant Mentions
**Action:**
  - Go through the conversation line by line.
  - Identify all first mentions of each restaurant and who made them.
  - Create a list of these initial mentions.

## Sub-task 2: Map Participants to Initial Suggestions
**Action:**
  - For each participant, list the restaurants they were the first to mention based on the extracted mentions.

## Sub-task 3: Construct the SuggestionTable
**Final Output Format:**
MentionedTable
- Rows: Participants extracted in Step 1.1.
- Columns: Restaurants extracted in Step 1.1.
- Fill each cell with "Mentioned" if the participant **first mentioned** the restaurant, or "None" if not.

---
**Note:**
- Complete each step thoroughly before moving to the next.
- Provide detailed findings at each step.
- Focus on who made the initial mentions of each restaurant.
- Adhere strictly to the output format.

**Output for Step 2:** <Mentioned Table>

**Prompt B.8 [Mixture of Reasoning Experts] for *Step 2***

# Step 2: Mention Analysis

Assume the perspective of each participant as an individual reasoning expert analyzing their own contributions to the restaurant mentions, specifically focusing on initial proposals.
**For each participant, perform the following analysis:**

---
## [Participant Name]'s Self-Analysis:

**Identify Personal Initial Mentions:**
  - Determine if you were the first to suggest any restaurant.
  - List the restaurants you initially proposed in the conversation.
  - Provide the context in which you made these initial mentions.

**Reflect on Others' Initial Mentions:**
  - Mention any reactions or support you gave to restaurants first mentioned by other participants.
  - Explain how your responses may have influenced the group's consideration of those restaurants.

---
After completing the analyses for all participants, integrate the findings:

**Action:**
  - Compile the individual contributions to create a comprehensive mapping of who first mentioned each restaurant.

**Final Output Format:**
MentionedTable
- Rows: Participants extracted in Step 1.1.
- Columns: Restaurants extracted in Step 1.1.
- Fill each cell with "Mentioned" if the participant **first mentioned** the restaurant, or "None" if not.

---
**Note:**
- Maintain each participant's perspective during each participant's self-analysis.
- Ensure the integration of findings reflects who first mentioned each restaurant.
- Adhere strictly to the final output format.

**Output for Step 2:** <Mentioned Table>

**Prompt B.9 [Chain-of-Thought] for *Step 3***

# Step 3: Perception Analysis (Emotion-Focused)

Using the results from Steps 1 and 2, perform a detailed analysis to determine the **emotional tone** of each participant's expression towards each restaurant mentioned in the conversation.
Focus exclusively on **emotional tone** (Positive, Negative, or Neutral) and do not consider external restrictions (e.g., restaurant availability or operational status) in your analysis.

**For each participant-restaurant pair:**
Review the conversation to identify the emotional tone (**Positive**, **Negative**, or **Neutral**) of all expressions made by the participants with respect to the restaurant.
Document changes in the emotional tone of the expressions of the participant, focusing on the emotional tone for the first and the final expression regarding the restaurant.

- Apply the following rules to decide what to record:

**Rules for Recording Emotional Tone:**
   1. **Consistent Emotional Tone:**
      - If the participant's emotional tone is always **Positive**, record **"Positive"**.
      - If the participant's emotional tone is always **Negative**, record **"Negative"**.
      - If the participant only expresses **Neutral** emotions or does not express any emotion toward the restaurant, record **"Neutral"**.

   2. **Change in Emotional Tone:**
      - If the participant record both **Positive** and **Negative** emotions towards the restaurant, record **Mix**.

**Conduct a comprehensive perception analysis:**
 - Step by step, document how you determined each participant's emotional tone for each restaurant.
 - Include quotations or references to specific parts of the conversation that support your perception analysis.

**Output Format:**
PerceptionTable
- Rows: Participants extracted in Step 1.1.
- Columns: Restaurants extracted in Step 1.1.
- Cell Values: One of the following, based on the participant's emotional tone, "Positive", "Negative", "Neutral", "Mix"

---
**Note:**
- Ensure that your analysis accurately captures any changes in emotional tone.
- Provide clear reasoning for each determination, focusing exclusively on emotional expressions, **excluding external restrictions or constraints**.
- Focus on being concise and precise in your explanations.
- Adhere strictly to the output format.

**Output for Step 3:** <Perception Table>

**Prompt B.10 [Self-Refinement] for** *Step 3*

# Step 3: Perception Analysis (Emotion-Focused)

Using the results from Steps 1 and 2, perform a detailed analysis to determine the **emotional tone** of each participant's expression towards each restaurant mentioned in the conversation. Focus exclusively on **emotional tone** (Positive, Negative, or Neutral) and do not consider external restrictions (e.g., restaurant availability or operational status) in your analysis.

**Initial Emotional Analysis:**
  - For each participant and restaurant, evaluate all expression regarding the restaurant and provide an initial assessment of changes in participant's emotional tone (e.g., includes both Positive and Negative emotions).
  - Determine the initial and final emotional tones.

**Self-Review of Emotional Analysis:**
  - Critically evaluate the initial emotional tone assessments for accuracy and completeness.
  - Ensure that changes in emotional tone are correctly identified according to the following rules:

**Rules for Recording Emotional Tones:**
   1. **Consistent Emotional Tone:**
     - If the participant's emotional responses are always **Positive**, record **"Positive"**.
     - If the participant's emotional responses are always **Negative**, record **"Negative"**.
     - If the participant only expresses **Neutral** emotions or does not express any emotion toward the restaurant, record **"Neutral"**.

   2. **Change in Emotional Tone:**
     -If the participant record both **Positive** and **Negative** emotions towards the restaurant, record **Mix**.

**Refined Emotional Analysis:**
  - Correct any errors or omissions identified during the self-review.
  - Present updated emotional assessments, ensuring that any changes in emotional tone are accurately represented with an asterisk (*).

**Final Output Format:**
PerceptionTable
- Rows: Participants extracted in Step 1.1.
- Columns: Restaurants extracted in Step 1.1.
- Cell Values: One of the following, based on the participant's emotional tone, "Positive", "Negative", "Neutral", "Mix"

---
**Note:**
- Separate the initial emotional analysis, self-review, and refined analysis clearly.
- Document any changes made during the refinement process.
- Ensure that emotional changes in perception are accurately captured and indicated.
- Focus on concise explanations to ensure consistency and clarity in evaluation.
- Adhere strictly to the final output format.

**Output for Step 3:** <Perception Table>

**Prompt B.11 [Prompt Decomposition] for** *Step 3*

# Step 3: Perception Analysis (Emotion-Focused)

Using the results from Steps 1 and 2, proceed through the following steps to determine the **emotional tone** of each participant's expression towards each restaurant mentioned in the conversation. Focus exclusively on **emotional tone** (Positive, Negative, or Neutral) and do not consider external restrictions (e.g., restaurant availability or operational status) in your analysis.

## Sub-task 1: Extract **Emotional Tone Expressions**
**Action:**
- Review the conversation to identify the emotional tone (**Positive**, **Negative**, or **Neutral**) of all expressions made by the participants with respect to the restaurant.
- For each participant-restaurant pair, document changes in the emotional tone of the expressions of the participant.
- Include direct quotes or references as evidence.

---
## Sub-task 2: Determine Final Emotional Tones**
**Action:**
- For each participant and restaurant, analyze changes in emotional tones:

**Rules for Recording Emotional Tones:**
  1. **Consistent Emotional Tone:**
     - If the participant's emotional responses are always **Positive**, record **"Positive"**.
     - If the participant's emotional responses are always **Negative**, record **"Negative"**.
     - If the participant only expresses **Neutral** emotions or does not express any emotional reaction toward the restaurant, record **"Neutral"**.

  2. **Change in Emotional Tone:**
     - If the participant record both **Positive** and **Negative** emotions towards the restaurant, record **Mix**.

---
## Sub-task 3: Construct the PerceptionTable**
**Output Format:**
PerceptionTable
- Rows: Participants extracted in Step 1.1.
- Columns: Restaurants extracted in Step 1.1.
- Cell Values: One of the following, based on the participant's emotional tone, "Positive", "Negative", "Neutral", "Mix"

---
**Note:**
- Complete each step thoroughly before moving to the next.
- Provide concise findings at each step, including the emotional tone sequences.
- Ensure that any emotional tone changes are accurately reflected in the final output.
- Adhere strictly to the output format.

**Output for Step 3:** <Perception Table>

**Prompt B.12 [Mixture of Reasoning Experts] for *Step 3***

# Step 3: Perception Analysis (Emotion-Focused)

Assume the perspective of each participant as an individual reasoning expert analyzing their **emotional tones** toward each restaurant, focusing exclusively on **emotional tone** and excluding any external restrictions or logistical considerations.
**For each participant:**

## [Participant Name]'s Emotional Tone Analysis:
**Identify Emotional Tones:**
- Review the conversation to identify the emotional tone (**Positive**, **Negative**, or **Neutral**) of all expressions made by the participants with respect to the restaurant.
- Document changes in the emotional tone of the expressions for each restaurant.

**Determine Final Emotional Tone and Changes in Emotional Tone:**
- For each restaurant, apply the following rules:

**Rules for Recording Emotional Tones:**
    1. **Consistent Emotional Tone:**
      - If the participant's emotional responses are always **Positive**, record **"Positive"**.
      - If the participant's emotional responses are always **Negative**, record **"Negative"**.
      - If the participant only expresses **Neutral** emotions or does not express any emotion toward the restaurant, record **"Neutral"**.

    2. **Change in Emotional Tone:**
      -If the participant record both **Positive** and **Negative** emotions towards the restaurant, record **Mix**.

**Provide Evidence:**
- Include quotes or references from the conversation that support the emotional tones and changes in emotional tone.

---
After completing the emotional analyses for all participants, integrate the findings:
**Action:**
- Compile the individual emotional tones to create a comprehensive mapping.
- Ensure that changes in emotional tone are accurately noted according to the rules defined earlier.

**Final Output Format:**
PerceptionTable
- Rows: Participants extracted in Step 1.1.
- Columns: Restaurants extracted in Step 1.1.
- Cell Values: One of the following, based on the participant's emotional tone, "Positive", "Negative", "Neutral", "Mix"

---
**Note:**
- Maintain each participant's emotional perspective during their self-analysis.
- Ensure the integration reflects individual emotional tones accurately, including any changes in emotional tone.
- Keep explanations concise to ensure fairness in evaluation.
- Adhere strictly to the final output format.

**Output for Step 3:** <Perception Table>

# Prompt B.13 [Chain-of-Thought] for *Step 4*

# Step 4: Perception Interpretation

Using the results from Steps 1 to 3, perform a detailed analysis to determine each participant's **preferences** and **constraints** regarding each restaurant mentioned in the conversation.

**For each participant and restaurant pair:**
- Review the conversation to identify any expressions of **preferences** or **constraints** made by the participant toward the restaurant.
- Assign the appropriate factors from the definitions provided, matching the participant's statements to the relevant factors.

# Factor Definitions:
　　**A1: Restaurant Quality**
　- Factors related to food quality, food genre, service quality, restaurant ambiance, seating capacity, and the overall quality of the restaurant.
　　**A2: Accessibility and Location**
　- The restaurant's access convenience, and the attractiveness of its surrounding area.
　　**A3: Schedule constraints**
　- The schedules of group members, the restaurant's business hours and operating days, and available reservation time slots.
　　**A4: Social Utility for Consensus**
　- The situation where individual member's preferences towards a particular alternative is affected by other members' preferences.
　　**A5: Inertia**
　- Prior experience with the restaurant, familiarity of members with the restaurant (i.e. frequent customers versus casual visitors), and the willingness or reluctance to explore new dining options (variety-seeking behavior).
　　**A6: Economic Considerations**
　- Price range, individual or group budget constraints, and cost-effectiveness.
　　**A7: Others**
　- Factors other than those mentioned above, or cases where no specific evidence was provided to support their expressions regarding the restaurants.

---
**Construct a comprehensive analysis:**
- Step by step, document how you determined each factor for each participant and restaurant pair.
- Include quotations or references to specific parts of the conversation that support your analysis.

---
**Output Formats:**
InterpretationTable
　- Rows: Participants extracted in Step 1.1.
　- Columns: Restaurants extracted in Step 1.1.
　- Cell Values: The factor codes (e.g., "A1", "A2") corresponding to the participant's preferences and constraints for that restaurant.
　　- If multiple factors apply, list them separated by commas.
　　- If no preference is expressed, enter **"None"** in the cell.

---
**Notes:**
- Ensure that your analysis accurately captures the participants' preferences and constraints.
- Provide clear reasoning for each determination.
- Adhere strictly to the output formats.

**Output for Step 4:** <Interpretation Table>

## Prompt B.14 [Self-Refinement] for *Step 4*

# Step 4: Perception Interpretation

Using the results from Steps 1 to 3, identify each participant's **preferences** and **constraints** regarding each restaurant mentioned.

**Initial Analysis:**
- For each participant and restaurant pair, provide your first assessment of any expressed **preferences** or **constraints**.
- Assign appropriate factor codes from the definitions provided.

# Factor Definitions:
   **A1: Restaurant Quality**
   - Factors related to food quality, food genre, service quality, restaurant ambiance, seating capacity, and the overall quality of the restaurant.
   **A2: Accessibility and Location**
   - The restaurant's access convenience, and the attractiveness of its surrounding area.
   **A3: Schedule constraints**
   - The schedules of group members, the restaurant's business hours and operating days, and available reservation time slots.
   **A4: Social Utility for Consensus**
   - The situation where individual member's preferences towards a particular alternative is affected by other members' preferences.
   **A5: Inertia**
   - Prior experience with the restaurant, familiarity of members with the restaurant (i.e. frequent customers versus casual visitors), and the willingness or reluctance to explore new dining options (variety-seeking behavior).
   **A6: Economic Considerations**
   - Price range, individual or group budget constraints, and cost-effectiveness.
   **A7: Others**
   - Factors other than those mentioned above, or cases where no specific evidence was provided to support their expressions regarding the restaurants.

**Self-Review:**
- Critically evaluate your initial assessments for accuracy and completeness.
- Ensure that you have correctly matched participants' statements to the relevant factors.
- Identify and correct any errors or omissions.

**Refined Analysis:**
- Present your updated assessments, reflecting any changes made during the self-review.

**Output Formats:**
InterpretationTable
  - Rows: Participants extracted in Step 1.1.
  - Columns: Restaurants extracted in Step 1.1.
  - Cell Values: The factor codes (e.g., "A1", "A2") corresponding to the participant's preferences and constraints for that restaurant.
    - If multiple factors apply, list them separated by commas.
    - If no preference is expressed, enter **"None"** in the cell.

**Notes:**
- Clearly separate the initial analysis, self-review, and refined analysis.
- Document any changes made during refinement.
- Adhere strictly to the final output formats.

**Output for Step 4:** <Interpretation Table>

**Prompt B.15 [Prompt Decomposition] for *Step 4***

# Step 4: Perception Interpretation

Proceed through the following steps to accurately determine each participant's **preferences** and **constraints** regarding each restaurant.

## Sub-task 1: Extract Preference and Constraint Expressions**
**Action:**
- Review the conversation to identify all expressions of **preferences** and **constraints** made by participants toward each restaurant.
- Document these expressions with direct quotes or references.

# Factor Definitions:
   **A1: Restaurant Quality**
   - Factors related to food quality, food genre, service quality, restaurant ambiance, seating capacity, and the overall quality of the restaurant.
   **A2: Accessibility and Location**
   - The restaurant's access convenience, and the attractiveness of its surrounding area.
   **A3: Schedule constraints**
   - The schedules of group members, the restaurant's business hours and operating days, and available reservation time slots.
   **A4: Social Utility for Consensus**
   - The situation where individual member's preferences towards a particular alternative is affected by other members' preferences.
   **A5: Inertia**
   - Prior experience with the restaurant, familiarity of members with the restaurant (i.e. frequent customers versus casual visitors), and the willingness or reluctance to explore new dining options (variety-seeking behavior).
   **A6: Economic Considerations**
   - Price range, individual or group budget constraints, and cost-effectiveness.
   **A7: Others**
   - Factors other than those mentioned above, or cases where no specific evidence was provided to support their expressions regarding the restaurants.

## Sub-task 2: Assign Factor Codes**
**Action:**
- For each expression identified, assign the appropriate factor code(s) from the definitions provided.
- Ensure that the factors accurately reflect the participants' statements.

## Sub-task 3: Construct the Tables**
InterpretationTable
  - Rows: Participants extracted in Step 1.1.
  - Columns: Restaurants extracted in Step 1.1.
  - Cell Values: The factor codes (e.g., "A1", "A2") corresponding to the participant's preferences and constraints for that restaurant.
    - If multiple factors apply, list them separated by commas.
    - If no preference is expressed, enter **"None"** in the cell.

**Notes:**
- Complete each step thoroughly before moving to the next.
- Provide concise findings at each step.
- Ensure that the factors are assigned accurately based on the participants' expressions.
- Adhere strictly to the output formats.

**Output for Step 4:** <Interpretation Table>

**Prompt B.16 [Mixture of Reasoning Experts] for *Step 4***

# Step 4: Perception Interpretation

Assume the perspective of each participant as an individual reasoning expert analyzing their **preferences** and **constraints** regarding each restaurant mentioned.
**For each participant:**

## [Participant Name]'s Analysis:
**Identify Preferences and Constraints:**
- Review the conversation to find all instances where you expressed **preferences** or **constraints** toward each restaurant.
- Include direct quotes or references from the conversation.

**Assign Factor Codes:**
- Match your expressions to the appropriate factor codes from the definitions provided.

# Factor Definitions:
   **A1: Restaurant Quality**
   - Factors related to food quality, food genre, service quality, restaurant ambiance, seating capacity, and the overall quality of the restaurant.
   **A2: Accessibility and Location**
   - The restaurant's access convenience, and the attractiveness of its surrounding area.
   **A3: Schedule constraints**
   - The schedules of group members, the restaurant's business hours and operating days, and available reservation time slots.
   **A4: Social Utility for Consensus**
   - The situation where individual member's preferences towards a particular alternative is affected by other members' preferences.
   **A5: Inertia**
   - Prior experience with the restaurant, familiarity of members with the restaurant (i.e. frequent customers versus casual visitors), and the willingness or reluctance to explore new dining options (variety-seeking behavior).
   **A6: Economic Considerations**
   - Price range, individual or group budget constraints, and cost-effectiveness.
   **A7: Others**
   - Factors other than those mentioned above, or cases where no specific evidence was provided to support their expressions regarding the restaurants.

---
After completing the analyses for all participants, integrate the findings:
**Action:**
- Compile the individual analyses to create the **InterpretationTable**.
InterpretationTable
  - Rows: Participants extracted in Step 1.1.
  - Columns: Restaurants extracted in Step 1.1.
  - Cell Values: The factor codes (e.g., "A1", "A2") corresponding to the participant's preferences and constraints for that restaurant.
    - If multiple factors apply, list them separated by commas.
    - If no preference is expressed, enter **"None"** in the cell.

**Notes:**
- Maintain each participant's perspective during their self-analysis.
- Ensure that the integration reflects the individual analyses accurately.
- Adhere strictly to the output formats.

**Output for Step 4:** <Interpretation Table>